\documentclass[letterpaper, 10 pt, conference]{ieeeconf}  

\IEEEoverridecommandlockouts                              

\pdfoutput=1 

\usepackage{times}
\usepackage[lcgreekalpha,notext]{stix}
\usepackage{epsfig}
\usepackage{graphicx}
\usepackage{amsmath}
\usepackage{amssymb}
\usepackage{algorithm}
\usepackage{algpseudocode}
\usepackage{tikz}
\usepackage[subrefformat=parens]{subcaption}
\usepgfmodule{plot}
\usepackage{pgfplots}
\usepgflibrary{arrows}
\usetikzlibrary{math}
\usepackage{pgffor} 
\usepgflibrary{shapes.geometric}
\usetikzlibrary{datavisualization}
\usepgflibrary{decorations.footprints}
\usetikzlibrary{intersections}
\usetikzlibrary{datavisualization.formats.functions}
\usetikzlibrary{plotmarks}
\usepgfplotslibrary{groupplots}
\usepgfplotslibrary{statistics}
\usepackage{mathtools}
\pgfplotsset{compat=1.9}
\usepackage[binary-units]{siunitx}
\usepackage{cite}
\usetikzlibrary{external}
\usepgfplotslibrary{external}
\definecolor{best}{rgb}{0,0.5,0.0}
\tikzstyle{format} = [draw, thin, fill=blue!20]
\tikzstyle{medium} = [ellipse, draw, thin, fill=green!20, minimum height=2.5em]
\tikzstyle{block} = [rectangle, draw, fill=blue!20, 
text width=5em, text centered, rounded corners, minimum height=4em]
\usepackage[export]{adjustbox}
\usepackage{diagbox} 
\pgfdeclarelayer{background}
\pgfdeclarelayer{foreground}
\pgfsetlayers{background,main,foreground}
\newcommand{\matr}{\mathbfit}
\renewcommand{\vec}{\mathbfit}
\newcommand{\norm}[1]{\left\lVert#1\right\rVert}
\newcommand{\Rot}{\matr{R}}
\newcommand{\tran}{\vec{t}}
\newcommand{\zer}{\matr{0}}
\newcommand{\Identity}{\matr{I}}
\newcommand{\Transformation}{\matr{T}}

\newcommand{\SE}{\mathrm{SE}(3)}
\newcommand{\Rdim}{\mathbb{R}}
\newcommand{\quat}{\vec{q}}
\newcommand{\chart}{\mathcal{C}}
\newcommand{\transpose}{\mathrm{T}}
\DeclareMathOperator*{\argmax}{arg\,max}

\usepackage{booktabs}
\title{\LARGE \bf Grasp Planning for Flexible Production with Small Lot Sizes based on CAD models using GPIS and Bayesian Optimization}

\author{Jianjie Lin, Markus Rickert and Alois Knoll
\thanks{Jianjie Lin, Markus Rickert and Alois Knoll  are with Robotics and Embedded System, Department of Informatics, Technische Universit\"at M\"unchen, Munich, Germany
        {\tt\small jianjie.lin@tum.de}, {\tt\small \{rickert, knoll\}@in.tum.de}}%
}

\begin{document}

\maketitle
\thispagestyle{empty}
\pagestyle{empty}

\begin{abstract}
Grasp planning for multi-fingered hands is still a challenging task due to the high nonlinear quality metrics, the high dimensionality of hand posture configuration, and complex object shapes. Analytical-based grasp planning algorithms formulate the grasping problem as a constraint optimization problem using advanced convex optimization solvers. However, these are not guaranteed to find a globally optimal solution. Data-driven based algorithms utilize machine learning algorithm frameworks to learn the grasp policy using enormous training data sets. This paper presents a new approach for grasp generation by formulating a global optimization problem with Bayesian optimization. Furthermore, we parameterize the object shape utilizing the Gaussian Process Implicit Surface~(GPIS) to integrate the object shape information into the optimization process. Moreover, a chart defined on the object surface is used to refine palm pose locally. 
We introduced a dual optimization stage to optimize the palm pose and contact points separately. We further extend the Bayesian optimization by utilizing the alternating direction method of multipliers~(ADMM) to eliminate contact optimization constraints. We conduct the experiments in the graspit! Simulator that demonstrates the effectiveness of this approach quantitatively and qualitatively. Our approach achieves a 95\% success rate on various commonly objects with diverse shapes, scales, and weights.
\end{abstract}
\section{Introduction}
\label{sec:introd}
\begin{figure}[t]
	\centering%
	\hfill%
	\begin{subfigure}[tb]{0.45\linewidth}%
		\centering%
		\includegraphics[width=\linewidth]{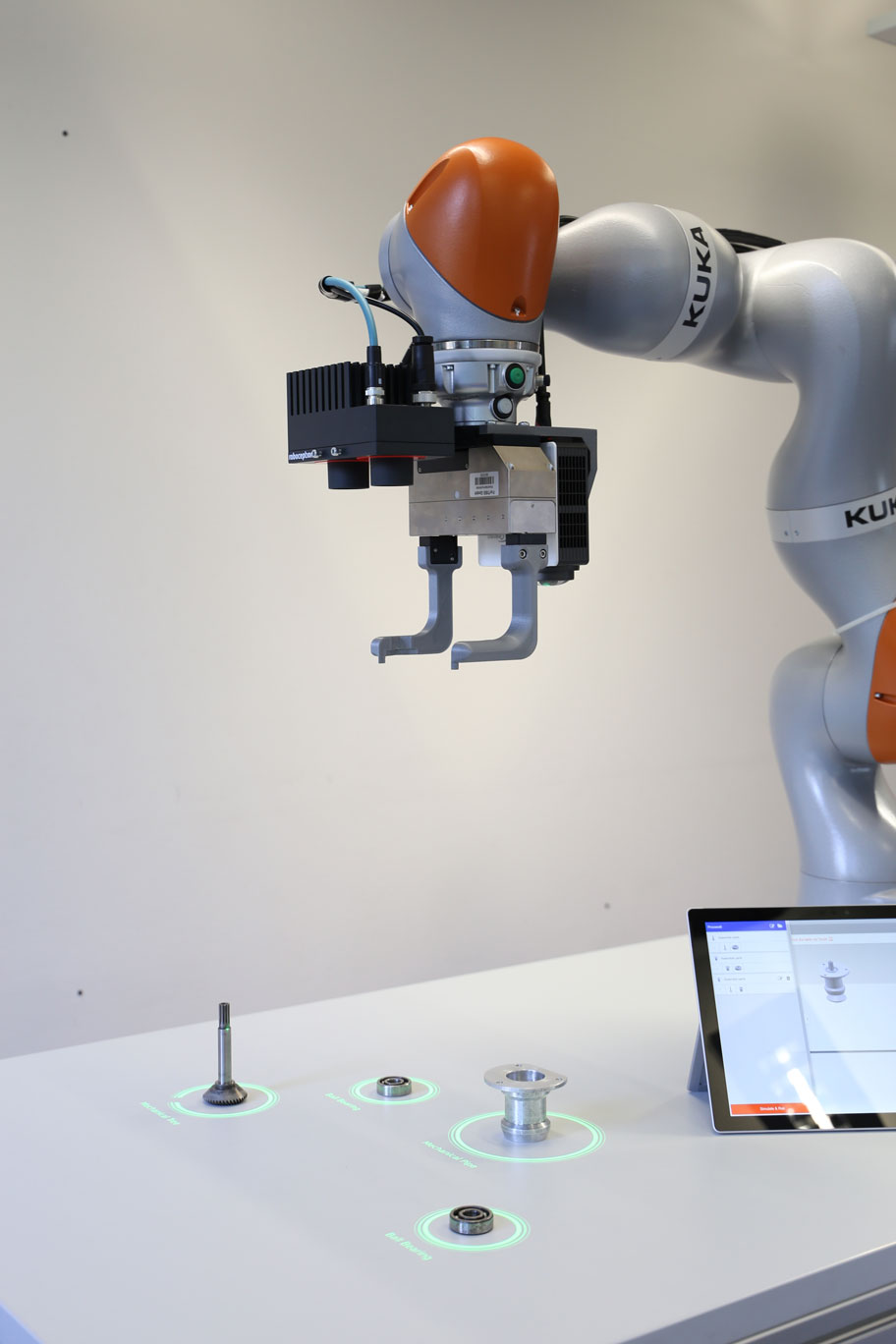}%
		\caption{}%
		\label{subfig:realrobotsetup}%
	\end{subfigure}%
	\hfill%
	\begin{subfigure}[tb]{0.45\linewidth}%
		\centering%
		\includegraphics[width=\linewidth,trim=585 0 585 0,clip]{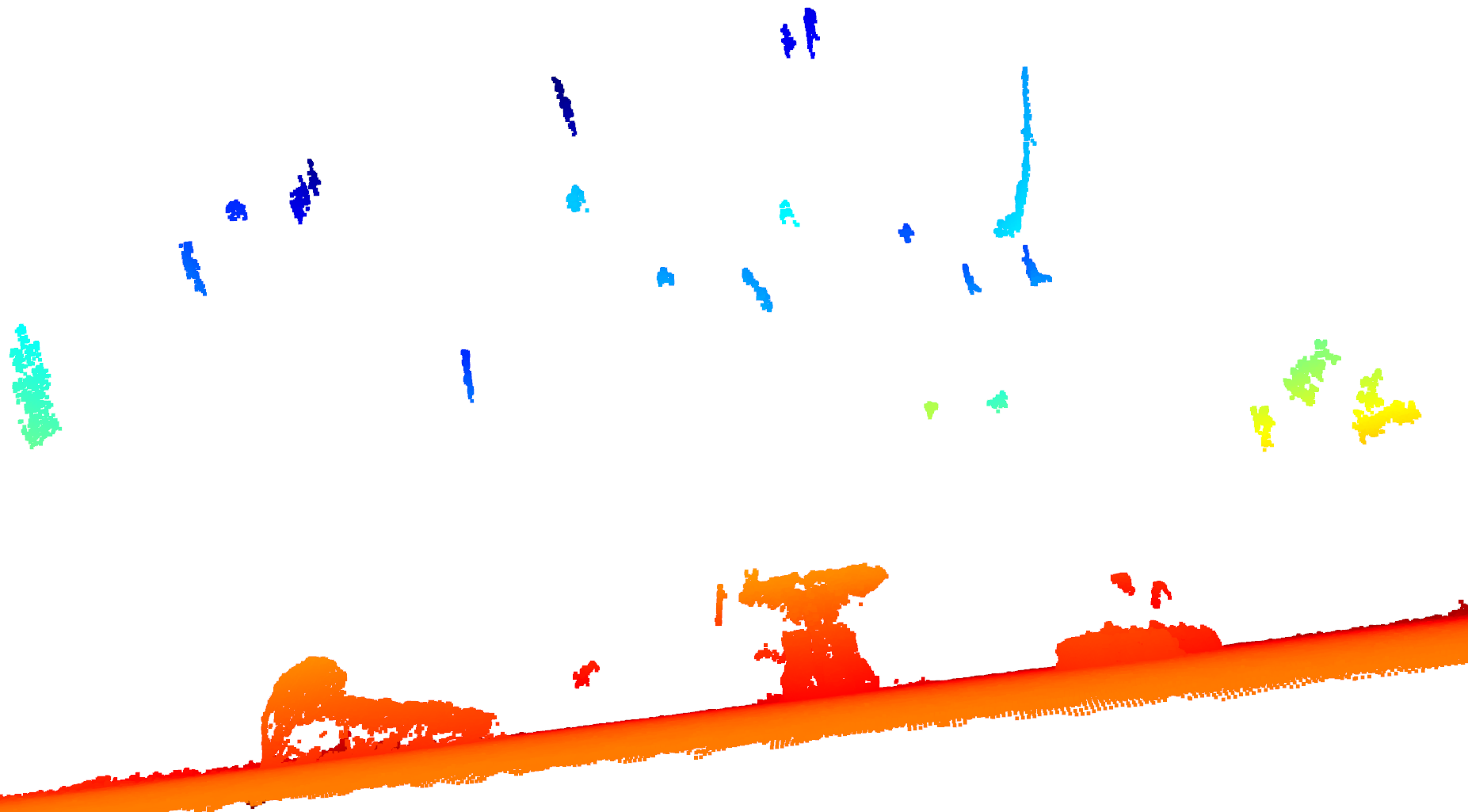}%
		\caption{}%
		\label{subfig:realsence}%
	\end{subfigure}%
	\hfill%
	\caption{Robot setup for assembling a gear box with \subref{subfig:realrobotsetup}~a lightweight robot and a 3D camera sensor, \subref{subfig:realsence}~point cloud scene of mechanical gearbox parts on the table.}
	\label{fig:robotsetup}
\end{figure}

Only a limited number of small and medium-sized enterprises in Europe use robot systems in production, mainly dealing with small lot sizes and requiring a more flexible production process. It is, however, very time-consuming and expensive to adapt a robot system to a new production line, and it requires expert knowledge for deploying such a system, which, however, is not commonly available in shop floor workers~\cite{SMErobotics2019}. Intuitive programming is currently proposed on the market for accelerating programming and remedying the problems caused by a lack of expert knowledge. Moreover, the service robots use semantic knowledge in combination with reasoning, and inference~\cite{Tenorth2012} to solve the declarative goal. Automatically synthesizing a robot program based on the semantic product, process, and resource descriptions enable an automatic adaptation to new processes. In this process, the recognition of objects and parts in the environments is involved, which is typically designed in CAD systems and described via a boundary representation~\cite{Perzylo2015b}. Due to the small lot size production of SMEs, it is not feasible to train the objects over a long period of time by using the data-driven approaches. Based on this observation, it will accelerate the deploying time if we grasp the object firstly in a simulator with the CAD models and then transfer the preplanned grasp to the real world with a 6D pose estimation~\cite{linGPIS2020}. Fig.~\ref{fig:robotsetup} shows an example of such a grasping use case for a mechanical gearbox together with a point cloud scene captured by the 3D camera sensor attached to the robot.

Dexterous robotic grasping planning has been an active research subject in the robotic community over the past decades. Grasping is essential in many areas such as industrial factories and household scenarios. There have many different kinds of robotic hands, i.e., traditional parallel-jaw grippers, complex multi-fingered hands, or even vacuum-based end effectors. The goal of grasp planning aims to find a proper contact on the object and an appropriate posture of the hand related to the object to maximize grasp quality. This is a challenging task, especially for multi-fingered hands, due to different kinds of object shapes, the complicated geometric relationship between robotic hands and objects, and the high dimensionality of hand configurations. 
Grasp planning can be divided into analytic approaches~\cite{bicchi2000robotic} on the one side and empirical or data-driven approaches on the other side~\cite{bohg2013data}. The analytic grasp synthesis approach is usually formulated as a constrained optimization problem over criteria that measure dexterity, equilibrium, and stability and exhibit a certain dynamic behavior. Besides, it requires the analysis of statically-indeterminate grasps~\cite{bicchi2000robotic} and under-actuated systems. The latter describes hands, in which the number of the controlled degrees of freedom is fewer than the number of contact forces, therefore further increasing the complexity of grasp synergies. One common assumption made in analytic methods is that precise geometric and physical models are available to the robot.
Furthermore, optimizing grasp quality with constraints based on a convex optimization solver such as SQP can not guarantee finding a good grasp. In contrast to analytic approaches, the empirical or data-driven approaches rely on sampling the grasp candidate either from a data set or by first learning a grasp quality and then selecting the best by ranking them according to some specific metric. In this work, we will study how to optimize the palm pose and contact point in the same framework by utilizing a global Bayesian optimization solver under consideration constraints. Gaussian Process Implicit Surface Atlas~(GPIS-Atlas) is used to parameterize the diverse shape. Therefore the geometry information can be integrated into the Bayesian optimization framework. Furthermore, GPIS-Atlas has the capability to describe the perfect geometry model in the form of a CAD model or the noisy point clouds \cite{williamsgaussian}. In the work~\cite{linGPIS2020}, the 6D pose is estimated between an object in the form of CAD model and the corresponding point clouds taken from an Ensenso Camera. Therefore, for finding an appropriate grasp pose for this object, we can directly apply our Bayesian optimization framework with the CAD model instead of working on noisy point clouds. After that, we transform the grasp pose using the 6D pose transformation from~\cite{linGPIS2020}.

\section{related work}
\label{sec:relatedWork}
Multi-fingered hand grasp planning is still challenging due to the high dimensionality of hand structure and complex graspable object shapes. Automatic grasp planning is a difficult problem because of the vast number of possible hand configurations. Several different approaches have been proposed to find an optimal grasp pose over the past decades. Goldfeder et al.~\cite{goldfeder2009columbia} introduced a database-backed grasp planning, which uses shape matching to identify known objects in a database with are likely to have similar grasp poses~\cite{goldfeder2009columbia}. Ciocarlie et al. presented~\cite{ciocarlie2010low} Eigengrasp planning defines a subspace of a given hand's DOF space and utilizes the Simulated Annealing planner to find an optimized grasp. Miller et al.~\cite{miller2003automatic} proposed a primitive shape-based grasp planning which generates a set of grasps by modeling an object as a set of shape primitives, such as spheres, cylinders, cones, and boxes. Pelosso et al.~\cite{pelossof2004svm} use an approach based on Support Vector Machines that approximate the grasp quality with a new set of grasp parameters. It considers grasp planning as a regression problem by given a feature vector, which should be defined heuristically. 
 With the continuous success of deep learning vision, researchers utilize deep learning, also in combination with reinforcement learning, to learn a grasp quality directly from an image via large training data sets~\cite{jang2017end}. Levine et al.~\cite{levine2018learning} used between 6 and 14 robots at any given point in time to collect data in two months and train a convolutional neural network to predict grasp success for a pick-and-place task with a parallel-jaw gripper. Mahler et al.~\cite{mahler2016dex} proposed a Dex-Net-based deep learning framework using a parallel-jaw gripper or vacuum-based end effector learn a grasp policy based on millions of grasp experiments. Kalashnikov et al.~\cite{kalashnikov2018qt} introduced a scalable self-supervised vision-based reinforcement learning framework to train a deep neural network Q-function by leveraging over 580k real-world grasp attempts. However, the deep learning-based algorithm can only take the 2d image as an input, and the trained neural network cannot be easily transferred to another robotic hand configuration. Varley et al. ~\cite{varley2017shape} obtained the geometry representation of grasping objects from point clouds using a 3D-CNN. Ten et.al~\cite{ten2017grasp} is the state of the art 6 DOF grasp planner~(GPD). Liang et.al.~\cite{liang2019pointnetgpd} proposed an end-to-end PointNetGPD to detect the grasp configuration from a point sets. Mousavian et al. introduced a 6DOF GraspNet by sampling a set of grasping using a variational autoencoder. In addition to deep learning, the Bayesian optimization-based algorithm in~\cite{nogueira2016unscented} can consider uncertainty in input space to find a safe grasp region by optimizing the grasp quality. Furthermore, it utilizes unscented transformation-based Bayesian optimization~(UBO), a popular nonlinear approximation method, to explore the safe region. However, UBO considers only the palm pose optimization without considering the contact point. 
We present a grasp planning approach in this work, where we combine Bayesian optimization with an analytical approach. We use the Grasp Wrench Space~(GWS)~\cite{ferrari1992planning} as grasp quality metric, which calculates the convex hull over discretized friction cones from the individual contact wrench spaces of all contacts. Due to the GWS metric's complexity, we explore the potential of Bayesian optimization to optimize this highly-nonlinear grasp quality problem. Since the hand posture~(hand palm pose) and hand configuration can be considered separately because the finger's contact points on the object surface only depend on the hand posture and forward kinematics of hand joints, we propose a dual-stage approach: In the first stage, we optimize the hand palm pose without considering hand configuration, and in the second stage, we use the result of the first stage and optimize the contact points on the object surface. Our approach optimizes a hand palm pose $\matr{T} \in \SE$ regarding its grasp quality. For this, we present the rotation in hyperspherical coordinates instead of a rotation matrix or quaternion. We further parameterize the object surface as a Gaussian Process Implicit Surface~(GPIS)~\cite{williamsgaussian} and use a $k$-D tree to find the closest point between the current palm pose and object surface. Based on GPIS, we can further compute a chart~$\chart$ and the corresponding normal vector~$\vec{N}_{\chart}$ on this nearest point. Utilizing this chart, we can make a local adaption of the palm pose to find a better location concerning the object surface. In the second stage, we convert the problem of solving constraints between the contact points and the object surface to querying the GPIS given a known contact point. Since the standard framework of Bayesian optimization cannot solve this constraint optimization problem, we use the Alternating Direction Method of Multipliers~(ADMM)~\cite{boyd2011distributed} to assist the contact point optimization by decomposing the whole problem into a set of subproblems.

\section{Problem Formulation}
\label{sec:problemFormulation}
In general, to define a grasp, we need two sets of variables: the intrinsic variables to define the hand degrees of freedom~(DOF) and the extrinsic variables to define the hand's position relative to the target object. Grasp planning is used to find the optimized contact points and an associated hand configuration to maximize grasp quality. The contact point on the object surface is denoted as $\vec{c}=[\vec{c}_1,\cdots,\vec{c}_n]$, where $\vec{c}_i\in \SE$, and~$n$ is the number of fingers. We will assume that contact happens on the fingertip, and one finger only has one contact on the object surface~$ \mathcal{O}$. The finger joint of the hand configuration is described as $\vec{q}=[\vec{q}_1,\cdots,\vec{q}_m]$, where~$m$ is the DOF. Note that some finger joints are under-actuated (passive joint). Therefore the number of finger joints is not equal to the DOF. The pose of the palm is represented by $\Transformation_{\mathrm{palm}}(\Rot,\tran)$. Mathematically, the optimized problem can be formulated as:
\begin{subequations}
    \begin{eqnarray}
    \label{eqn:basicProblema}
    &\underset{\vec{c},\vec{q},\Transformation_{\mathrm{palm}}}\max Q(\vec{c},\vec{q},\Transformation_{\mathrm{palm}})\\
    \label{eqn:basicProblemb}
    \text{s.t.}& \vec{c}=\operatorname{FK}_{\mathrm{palm}2\mathrm{c}}(\vec{q},\Transformation_{\mathrm{palm}}) \in \mathcal{O}\\
    \label{eqn:basicProblemd}
    & q_{\mathrm{min},i}\leq q_i \leq q_{\mathrm{max},i},& i=1\cdots m \, ,
    \end{eqnarray}
    \label{eqn:basicProblem}
\end{subequations}
where $ Q(\vec{c},\vec{q}, \Transformation_{\mathrm{palm}})$ is the GWS, which is a 6-dimensional convex polyhedron, the epsilon and volume quality metric introduced by Ferrari and Canny~\cite{ferrari1992planning}. The epsilon quality is defined as the minimum distance from the origin to any of the hyperplanes
defining the boundary of the GWS. In contrast, the volume quality is the volume of GWS. $\operatorname{FK}_{\mathrm{palm}2\mathrm{c}}$ is the forward kinematics from the palm pose to the contact points. The formulation~(\ref{eqn:basicProblemb}) constrains all contact points on the object surface~$ \mathcal{O}$. Furthermore, a contact is defined as any point where two bodies are separated by less than the contact threshold~$\epsilon_c$, but not interpenetrating. In this work, the object surface will be parameterized by using GPIS to easily check if the contact point satisfies the constraint~(\ref{eqn:basicProblemb}). The problem in~(\ref{eqn:basicProblem}) is a high-dimensional nonlinear constraint problem, besides the gradient of the objective function and constraints cannot be analytically computed. Further, the convex optimization solver can only find a local minimum. Based on those observations, Bayesian optimization is applied to find a near-global optimization solution for these problems. Since the palm pose and contact point are nearly independent, we can switch between optimizing the palm pose and the contact point.

\section{Bayesian Optimization for Grasp Planning}
\label{sec:bo}
Bayesian optimization is a global optimization method, which can be used to solve the problem 
\begin{align}
x_{\mathrm{optimized}}=\underset{x \in \mathcal{X}}\max f_{\mathrm{obj}}(x) \, ,
\end{align}
where the objective function~$f_{\mathrm{obj}}(x)$ is a black-box function or a function which is expensive to evaluate and \hbox{$\mathcal{X} \subseteq \Rdim^D$} is a bounded domain.
We use the Latin Hypercube Sampling~(LHS) to get the initial sampling, and save it as data set~$ \mathcal{D}_{0:t-1}=\bigl\lbrace\left(\mathbf{x}_{0}, y_{0}\right), \ldots,\left(\mathbf{x}_{t-1}, y_{t-1}\right)\bigr\rbrace$, and learn a Gaussian process model~(GP). The essential step is to choose an appropriate acquisition function. Here, we use Expected Improvement~\cite{Jones1998}, which is defined as
\begin{align}
\operatorname{EI}(\vec{x})=\mathbb{E}\bigl[\max \left(f_{\mathrm{obj}}(\vec{x})-f_{\mathrm{obj}}(\vec{x}^{+}), 0\right)\bigr] \, ,
\end{align}
where $\mathbb{E}$ is the expectation function, $f_{\mathrm{obj}}(\vec{x}^{+})$ is the best observation with the location $\vec{x}^{+}$ so far. The $ \operatorname{EI}(\vec{x})$ can be evaluated analytically under the GP model as \[ \operatorname{EI}(\vec{x})= \nonumber \mathbb{1}\bigl(\sigma(\vec{x})\bigr)\Bigl(\bigl(\mu(\vec{x})-f_{\mathrm{obj}}(\vec{x}^{+})-\xi\bigr) \Phi(Z)+\sigma(\vec{x}) \phi(Z)\Bigr) \, , \]
with $ Z=\mathbb{1}\bigl(\sigma(\vec{x})\bigr)\Bigl(\frac{\mu(\vec{x})-f_{\mathrm{obj}}(\vec{x}^{+})-\xi}{\sigma(\vec{x})}\Bigr)$,
where $\mathbb{1}(x)$ is the indicator function that is equal to~$0$ for $x \leq 0$ and equal to~$1$ otherwise. The mean~$\mu(\vec{x})$ and standard deviation~$\sigma(\vec{x})$ are defined in the GP posterior at~$\vec{x}$, and~$\Phi$ and $\phi$ are the CDF~(cumulative distribution function) and PDF~(probability density function) of the standard normal distribution. $\xi$ is a parameter which balances between the exploration and exploitation. The objective function in our algorithm for optimizing the grasp contact points is grasp quality which consist of epsilon and volume quality.
\begin{align}
f_{\mathrm{obj}}(\vec{x})=\bigl(\mathbb{1}(q_{\epsilon})q_{\epsilon}+\lambda \, q_{\mathrm{volume}}\bigr) \, ,
\end{align}
where $\lambda$ is a predefined parameter. The Mat\'ern covariance function~($\nu=5/2$) is chosen as kernel function for the Gaussian process model in the Bayesian optimization and can be described as: $K_{5 / 2}(d)=\sigma^{2}\left(1+\frac{\sqrt{5} d}{\rho}+\frac{5 d^{2}}{3 \rho^{2}}\right) \exp \left(-\frac{\sqrt{5} d}{\rho}\right)$ with hyper parameter~$\sigma$ and length-scale~$\rho$. The parameter~$d$ is the distance between two query points. Since we need to optimize the palm pose, which is interpreted as a transformation. We define the distance between two transformation matrices as
~$ \Delta_{\Transformation}=\norm{\tran_1-\tran_2}+\norm{\operatorname{log}(\Rot_1^\transpose\Rot_2)}_{\mathrm{F}}/\sqrt{2}$, 
where the term~$ \norm{\operatorname{log}(\Rot_1^\transpose\Rot_2)}_{\mathrm{F}}/\sqrt{2}$ is the geodesic distance defined in the Riemann manifold.
To optimize the hyperparameter, the algorithm RProp~(Resilient Propagation)~\cite{blum2013optimization} is applied, a popular gradient descent algorithm that only uses the signs of gradients to compute updates and can dynamically adapt the step size for each weight independently.

The object surface in our algorithm is described as a GPIS and every point lying on the surface in the set $ \mathcal{X}_{\prime}$ should satisfy the equality constraints~\hbox{$\mathcal{X}_{\prime}=\bigl\lbrace\mathbf{x} \in \mathbb{R}^{3} : f_{\mathrm{GPIS}}(\vec{x})=0\bigr\rbrace$}. Furthermore, the tangent space of each point on these surface will be computed by using
\begin{align}
\begin{bmatrix}{\nabla f_{\mathrm{GPIS}}^{\transpose}(\vec{x}_i)} \\[0.5em] {\matr{\Phi}_{i}^{\transpose}(\vec{x}_i)}\end{bmatrix} \matr{\Phi}_{i}(\vec{x}_i)=\begin{bmatrix}{\zer} \\ {\Identity}\end{bmatrix} \, ,
\label{eqn:tangentSpace}
\end{align}
where $\matr{\Phi}_{i}(\vec{x}_i) \in \Rdim^{3\times 2}$ is the basis of tangent space at the location $\vec{x}$ and $ \nabla f_{\mathrm{GPIS}}(\vec{x}_i)\in \Rdim^{3\times 1}$ is the gradient of implicit function for $\vec{x}$. By using $ \matr{\Phi}_{i}(\vec{x})$, we can map~$\vec{x}_i$ to~$\vec{x}_i^{\prime}$ with~$ \vec{x}_{i}^{\prime}=\vec{x}_{i}+\matr{\Phi}_{i}(\vec{x}_i) \, \vec{u}_{i}$,
where $\vec{u}_i \in \Rdim^{2\times 1}$ is a point in the local coordinate on this chart.
\begin{figure}[b]
	\centering
	\begin{subfigure}[b]{0.9\linewidth}%
		\includegraphics[width=1\textwidth]{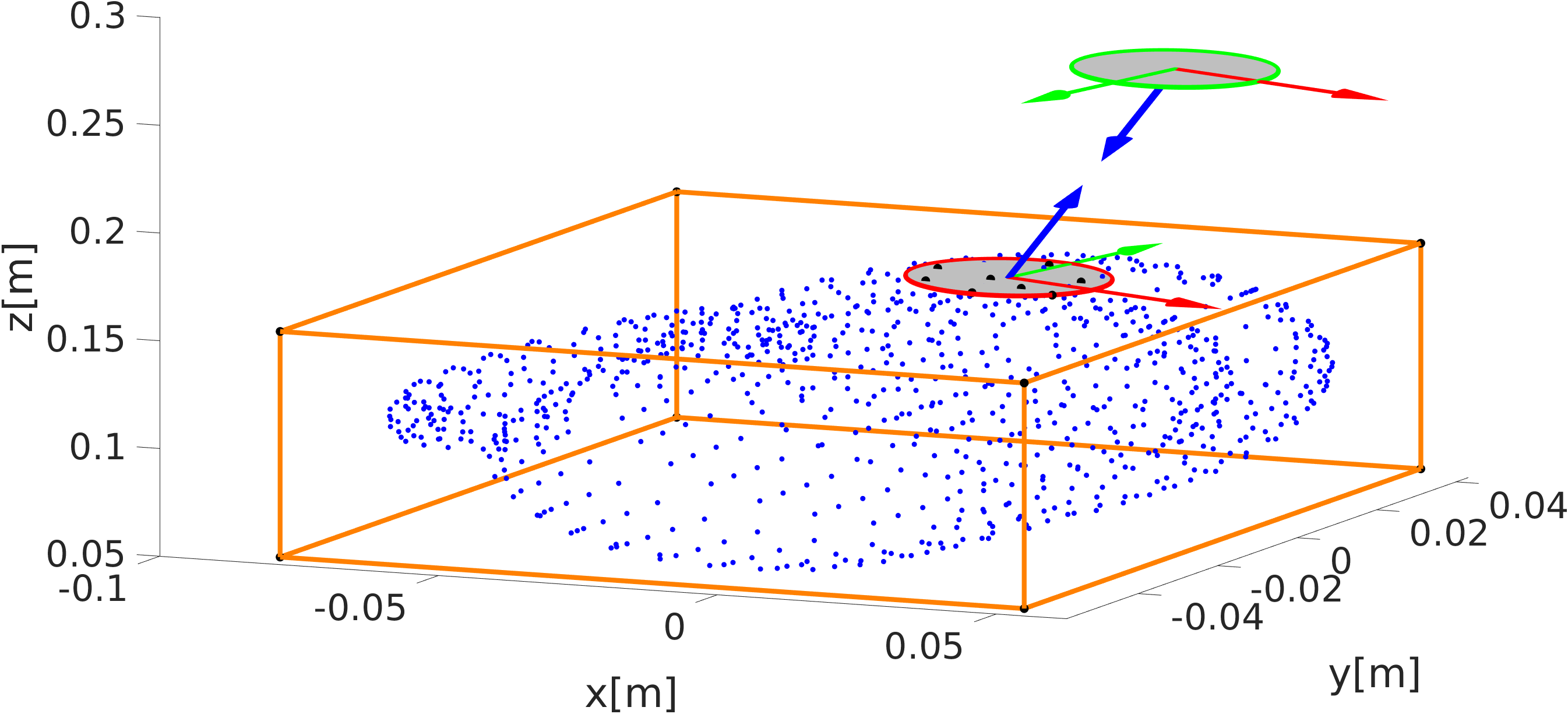}%
		\caption{}%
		\label{fig:Chart}%
	\end{subfigure}%
	\hfill%
	\begin{subfigure}[b]{0.9\linewidth}%
		\includegraphics[width=1\textwidth]{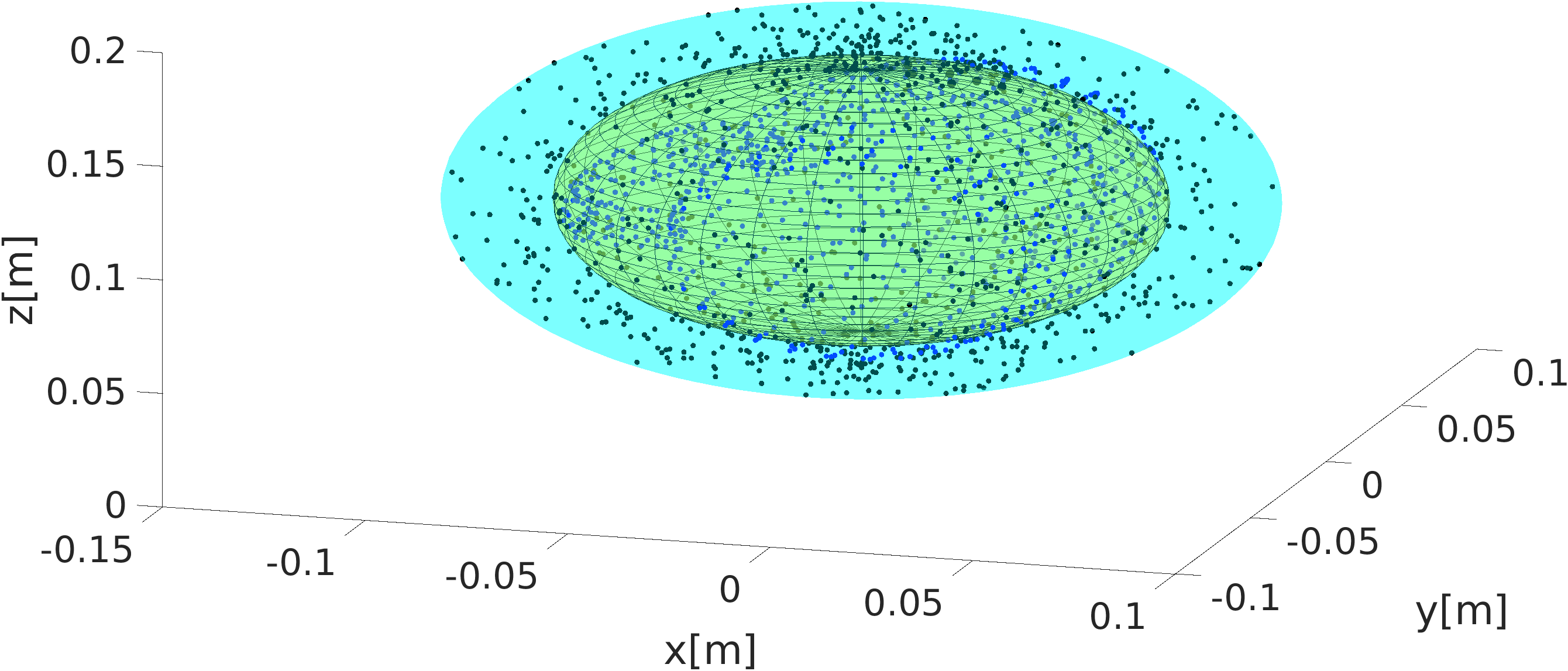}%
		\caption{}%
		\label{fig:MugSampling}%
	\end{subfigure}%
	\caption{Visualization of \subref{fig:Chart}~Tangent space on a mug surface and \subref{fig:MugSampling}~sampling for initial palm pose.}%
	\label{fig:ChartMugSampling}
\end{figure}
The sample value~$\vec{x}_i^{\prime}$ is shown as black dots in Fig.~\ref{fig:Chart}, the tangent vector on the chart is shown as red and green arrows, while the blue arrow shows the normal vector.
\subsection{Hand Palm Pose Optimization~(HPP-Opt)}
To optimize the palm pose, we need to take the transformation~\hbox{$\Transformation_{\mathrm{palm}}(\Rot,\tran)\in \SE$} into account, where the rotation matrix~$\Rot$ can be interpreted by using a unit quaternion~$\quat$. Since the unit quaternion manifold~$\mathcal{M}_{\mathbb{H}}$ is an Riemannian manifold, by virtual equality of $ \mathcal{M}_{\mathbb{H}}$ and 4D unit hypersphere~$\mathcal{S}^{3}=\bigl\lbrace x \in \mathbb{R}^{3+1} :\norm{\vec{x}}=1\bigr\rbrace$, the quaternion~$\quat$ can be represented in the hyperspherical coordinates with $\phi,\psi,\theta$ where $\phi,\psi$ range over~$[0,\pi]$ and $\theta$ ranges over $[0,2\pi)$. 
Employing hyperspherical coordinates, the constraints of $\quat$ disappear. Therefore, the palm pose optimization is converted to an unconstrained optimization, and it can be mathematically formulated as
\begin{align}
	\label{eqn:palmposea}
	\underset{\phi,\psi,\theta,\vec{t}_{\mathrm{palm}}}\max  f_{\mathrm{obj}}\bigl(\vec{q},\Transformation_{\mathrm{palm}}(\Rot(\phi,\psi,\theta),\vec{t}_{\mathrm{palm}})\bigr) \, .
\end{align}
To reduce the search space, we constrain the palm pose between two bounding boxes, represented by an axis-aligned minimum bounding box~(AABB), denoted as $\mathcal{X}_{\mathrm{AABB},i}$ with $i \in \{1,\, 2\}$. The smaller one is shown as an orange cube in Fig.~\ref{fig:Chart}. We define the variable $\vec{x}_{\mathrm{palm}}=[\tran,\phi,\psi,\theta]^\transpose\in\Rdim^{6\times1}$.
Sampling a point between two bounding box cannot be formulated mathematically, therefore, we use an ellipsoid~$\mathcal{E}_1=(a_1,b_1,c_1,a_0,b_0,c_0)$ to approximate $\mathcal{X}_{\mathrm{AABB},1}$ and another ellipsoid~$\mathcal{E}_2=(a_2,b_2,c_2,a_0,b_0,c_0)$ to approximate the bigger bounding box~$\mathcal{X}_{\mathrm{AABB},2}$. As the result, the palm pose sample~$\vec{t}=[t_x,t_y,t_z]$ can be formulated as
\begin{align*}
t_x&=a_0+r*a_1\sin(\theta)\cos(\phi)\\
t_y&=b_0+r*b_1\sin(\theta)\sin(\phi)\\
t_z&=c_0+r*c_1\cos(\theta) \, ,
\end{align*}
where $m_0=\frac{k_{\mathrm{min},1}+k_{\mathrm{max},1}}{2}$, $m_1=\frac{k_{\mathrm{max},1}-k_{\mathrm{min},1}}{2}$, \hbox{$m\in\{a,b,c\}$}, and~$k \in \{x,y,z\}$. The parameter~$r$ is a random variable which ranges over~$[1,r_{\max}]$, where
\[r_{\mathrm{max}}=\sqrt{\frac{1}{\bigl(\frac{a_1\sin(\theta)\cos(\phi)}{a_2}\bigr)^2+\bigl(\frac{b_1\sin(\theta)\sin(\phi)}{b_2}\bigr)^2+\bigl(\frac{c_1\cos(\theta)}{c_2}\bigr)^2}} \, . \]
The parameter $\theta$ ranges over~$[0,\pi]$
and $\theta$ ranges over~$[0,2\pi)$. The two ellipsoids are shown in Fig~\ref{fig:MugSampling}.
The problem in~(\ref{eqn:palmposea}) can be solved by using Bayesian optimization, as shown in Alg.~\ref{alg:Palm Pose Optimization}.
The solution found by Bayesian optimization is based on the probability of best grasp distribution. To improve the performance, a local adaption based on GPIS-Atlas is proposed.

\subsubsection{GPIS-Atlas based local adaption} 
The first step is to find the closest point from the current palm pose to the object using $k$-D tree algorithm. Assuming we found the pose~$ \vec{t}_{\mathrm{closest}}$, a chart~$\mathcal{C}_i$ with the center point $\vec{t}_{\mathrm{closest}}$ is created by solving~(\ref{eqn:tangentSpace}). We denoted the outward unit normal vector of this chart~$\mathcal{C}_i$ as $\vec{N}_{\mathrm{closest}}$ and the robot hand posture is designed to orient to the direction of a chart normal $\vec{N}_{\chart}$. We apply the following approaches to get the pose~$\Transformation$. Assuming that the normal vector of hand in the initial state points to the z-axis~$\vec{n}_z$, the hand is currently in local frame~$\mathcal{H}_1$ with rotation matrix $ {}^\mathcal{W}\Rot_{\mathcal{H}_{1}}$ with respect to the world frame $\mathcal{W}$. The next hand configuration should point to the normal direction~$\vec{N}_{\chart}$ in local frame~$\mathcal{H}_{\mathcal{C}_i}$, therefore the corresponding rotation matrix $ {}^\mathcal{W}\Rot_{\mathcal{H}_{\mathcal{C}_i}}$ can be interpreted in angle-axis representation $[n_{\mathrm{axis}},\theta_{\mathrm{axis}}]$ with $\theta_{\mathrm{axis}}=\operatorname{atan2}(\norm{\vec{n}_z\times\vec{N}_{\chart}},\vec{n}_z\cdot\vec{N}_{\chart})$ and $ n_{\mathrm{axis}}=\vec{n}_z\times\vec{N}_{\chart}$. As a result, we can transform the local frame~$ \mathcal{H}_1$ to ~$ \mathcal{H}_{\mathcal{C}_i}$ with the rotation transformation as $ {}^{\mathcal{H}_1}\Rot_{\mathcal{H}_{\mathcal{C}_i}}={}^\mathcal{W}\Rot_{\mathcal{H}_{1}}^\transpose{}^\mathcal{W}\Rot_{\mathcal{H}_{\mathcal{C}_i}}$. Furthermore, the translation of the palm pose is defined as $\vec{t}_{\mathrm{palm}}=\vec{\vec{t}_{\mathrm{closest}}}+\lambda\vec{N}_{\mathrm{closest}}$. This means that the new palm pose~$\vec{x}_{\mathrm{palm}}$ is parallel to the chart~$\mathcal{C}_i$ with the distance~$\norm{\lambda}$, as shown in Fig.~\ref{fig:Chart}. The parameter is optimized so that the hand is not colliding with the object. The whole transformation is defined as $\Transformation_{\mathrm{palm}}=\Transformation(\Identity,\vec{t}_{\mathrm{palm}}) \, \Transformation({}^\mathcal{W}\Rot_{\mathcal{H}_{\mathcal{C}_i}},\zer) \, \Transformation(\Rot_z,\zer)$. The transformation~$\Transformation(\Rot_z,\zer)$ is used to further guarantee no collision. Furthermore, we can define a point set on the chart~$\mathcal{C}_i$ as $ \mathcal{X}_{\mathcal{C}_i}$ and randomly choose a sample~$ \vec{t}_{\mathrm{sample}}\in \mathcal{X}_{\mathcal{C}_i}$ as $ \vec{t}_{\mathrm{closest}}$. 
\begin{algorithm}[t]
    \small
    \begin{algorithmic}[1]
        \Require $n_{\mathrm{dim}}$, $n_{\mathrm{iter}}$ 
        \State Get LHS $\mathcal{D}_{0:t-1}=\bigl\lbrace\left(\mathbf{x}_{0}, y_{0}\right), \ldots,\left(\mathbf{x}_{t-1}, y_{t-1}\right)\bigr\rbrace$
        \State Fit the Gaussian process Model $p\bigl(y | \vec{x}, \mathcal{D}_{0:t-1}\bigr)$
        \State Optimized hyper Parameter Rrop
        \For {$\mathrm{t}=1$ to $n_{\mathrm{iter}}$} 
        \State $\mathbf{x}_{\mathrm{palm},t}=\underset{\vec{x}}\argmax \operatorname{EI}\left(\mathbf{x} | \mathcal{D}_{0 : t-1}\right)$
        \State Find a collision-free contact and get $\vec{q},\vec{x}_{\mathrm{palm}}$
        \State $\mathcal{G}=\operatorname{GPISAtlas}(f_{\mathrm{GPIS}}, \vec{x}_{\mathrm{palm}})$
        \State $\bigl\lbrace\mathcal{G}_{\mathrm{best}}(\vec{x}_{\mathrm{palm},t},\vec{q}),y_{\mathrm{max},t}\bigr\rbrace=\underset{\mathcal{G}}\argmax f_{\mathrm{obj}}(\vec{q},\vec{x}_{\mathrm{palm}})$
        \State Add the new sample $\mathcal{D}_{0 : t}=\bigl\lbrace\mathcal{D}_{0 : t-1},(\vec{x}_{\mathrm{palm},t},y_{\mathrm{max},t})\bigr\rbrace$
        \State update Gaussian process model~(GP)
        \EndFor
        \State $\vec{x}_{\mathrm{palm}}^{+}=\underset{\vec{x}_{i} \in \vec{x}_{0 : t}}\argmax f_{\mathrm{obj}}\left(\vec{x}_{i}\right)$
        \State \Return $\vec{x}_{\mathrm{palm}}^{+}$
    \end{algorithmic} 
    \caption{Palm Pose Optimization~{HPP-Opt}}
    \label{alg:Palm Pose Optimization}
\end{algorithm}
\begin{algorithm}[t]
    \small
    \begin{algorithmic}[1]
        \Require $f_{\mathrm{GPIS}}$, $\vec{x}_{\mathrm{palm,current}}$
        \State Set $\vec{t}=\vec{x}_{\mathrm{palm,current}}(0:2)$
        \State Get $\vec{t}_{\mathrm{closest}}=\operatorname{KD}(\vec{t}_{\mathrm{palm,current}})$
        \State Set $\mathcal{G}=\{\}$, $\lambda \leftarrow 0$, $\theta_z \leftarrow 0$, $\lambda_{\mathrm{max}}$, $n_{\mathrm{seed}}$
        \State Compute Chart $\mathcal{C}_i$ and $\vec{N}_{\mathrm{closest}}$ by equation~\ref{eqn:tangentSpace}
        \For{n = 0 to $n_{\mathrm{seed}}$ }
        \State $\vec{t}_{\mathrm{closest}}^{\prime}=\vec{t}_{\mathrm{closest}}+\matr{\Phi}_{i}(\vec{t}_{\mathrm{closest}}) \mathbf{u}_{\mathrm{rand}}$
        \While {\text{CheckCollsion}}
        \State $\vec{t}_{\mathrm{palm}}=\vec{t}_{\mathrm{closest}}^{\prime}+\lambda\vec{N}_{\mathrm{closest}}$
        \If{$\lambda<\lambda_{\mathrm{max}}$}
        \State $\Transformation_{\mathrm{palm}}=\Transformation_1(\Identity,\vec{t}_{\mathrm{palm}})\Transformation_2({}^\mathcal{W}\Rot_{\mathcal{H}_{\mathcal{C}_i}},\zer)$ 
        \State $\lambda\leftarrow\lambda+\Delta_\mathrm{\lambda}$
        \Else
        \State $\Rot_z=\mathrm{Angleaxis}(0,0,1,\theta_z)$
        \State $\theta_z\leftarrow \theta_z+\Delta_{\theta_z}$
        \State $\Transformation_{\mathrm{palm}}=\Transformation_1(\Identity,\vec{t}_{\mathrm{palm}})\Transformation_2({}^\mathcal{W}\Rot_{\mathcal{H}_{\mathcal{C}_i}},\zer)\Transformation_3(\Rot_z,\zer)$ 
        \EndIf
        \EndWhile
        \State Execute the Grasp, get hand configuration $\vec{q}$
        \State $\mathcal{G}=\mathcal{G}\cup\{\vec{x}_{\mathrm{palm}},\vec{q}\}$
        \EndFor
        \State \Return $\mathcal{G}$
    \end{algorithmic} 
    \caption{GPISAtlas() function for local adaption}
    \label{alg:GPISAtlasbasedlocal adaption}
\end{algorithm}

\subsection{ADMM-Based Contact Point Optimization~(ADMM-CP-Opt)}

The contact point optimization is used to find a set of desired joints~$\vec{q}$ for each finger and the contact points~$\vec{c}$ on the object surface. It can be described as
\begin{subequations}%
	\begin{eqnarray}%
	& \underset{\vec{c}, \vec{q}}\max&f_{\mathrm{obj}}(\vec{q},\Transformation_{\mathrm{palm}})\\
	&\text{s.t.}&\vec{c}=\operatorname{FK}(\vec{q},\Transformation_{\mathrm{palm}}),\\
	& & \lvert f_{\mathrm{GPIS}}(\vec{c}_k) \rvert \leq\epsilon_c,\qquad k=1\cdots n\\
	& & q_k \in [q_{\mathrm{min},k}, q_{\mathrm{max},k}],\qquad k=1\cdots m \, ,
	\end{eqnarray}\label{eqn:contactPose}%
\end{subequations}
where $\operatorname{FK}$ calculates the forward kinematics, and~$n$ represents the number of contact points on the object surface. The parameter~$m$ is the DOF of a hand. A contact point is represented as a transformation~$\Transformation_{c_i}$. However, each finger has fewer joints than $6$, which results in an underestimated system. Consequently, we cannot directly calculate the inverse kinematics based on the contact points, and it is not possible to arbitrarily move the fingertip on the object surface. To relax the constraints, we will not fix the palm pose, but constrain the palm pose on the chart~$\chart_{\mathrm{palm},i}$, which is parallel to chart~$\chart_i$ on the object surface, therefore \hbox{$\Transformation_{\mathrm{palm}}\in \chart_{\mathrm{palm},i}$}. Since the equality constraints cannot be solved by using Bayesian optimization, the Alternating Direction Method of Multipliers~(ADMM) based Bayesian optimization~\cite{boyd2011distributed} is utilized to solve the contact pose optimization problem~(\ref{eqn:contactPose}) with the new formulation
\begin{align} 
\underset{\vec{q}\in \vec{B}}\max f_{\mathrm{obj}}(\vec{q},\Transformation_{\mathrm{palm}})+g_c(\vec{q,\Transformation_{\mathrm{palm}}}) \, ,
\label{eqn:unconstraintedOpt}
\end{align}
with $g_c(\vec{q, \Transformation_{\mathrm{palm}}})= \mu\sum_{i=0}^n\vec{c}_i(\vec{q},\Transformation_{\mathrm{palm}} )^2$ and  $\vec{c}_i(\vec{q},\Transformation_{\mathrm{palm}})=\biggl\lvert f_{\mathrm{GPIS}}\Bigl(\operatorname{FK}_i \bigl(\vec{q},\Transformation_{\mathrm{palm}}(\Rot,\tran)\bigr)\Bigr)\biggr\rvert-\epsilon_c$. In order to solve~(\ref{eqn:unconstraintedOpt}), ADMM introduces an auxiliary variable~$\vec{z}$, resulting in
\begin{subequations}%
    \begin{eqnarray}%
   & \underset{\vec{q},\vec{z} \in \vec{B}}\max&f_{\mathrm{obj}}(\vec{q},\Transformation_{\mathrm{palm}} )+\operatorname{g}_c(\vec{z},\Transformation_{\mathrm{palm}})\\
   &\text{s.t.}& \vec{q}=\vec{z} \, .
    \end{eqnarray}%
    \label{eqn:admm}%
\end{subequations}
In the following, we neglect~$\Transformation_{\mathrm{palm}}$ in~$f_{\mathrm{obj}}$ and~$g_c$. By applying Augmented Lagrangian function for equation~(\ref{eqn:admm}), a new objective function is formulated as
\begin{align}
\mathcal{L}_{\rho}(\vec{q}, \vec{z}, \vec{y})=f_{\mathrm{obj}}(\vec{q})+g_c(\vec{z})+\frac{\rho}{2}\norm{\vec{q}-\vec{z}+\frac{\vec{y}}{\rho}}_{2}^{2}
\end{align}
Therefore, we can solve~$f_{\mathrm{obj}}(\vec{q})$ and~$g_c(\vec{z})$ by alternating over the following sub problems:
 \begin{subequations}%
     \begin{align}%
     \label{eqn:admma}
     \vec{q}^{k+1}&=\underset{\vec{q}}\argmax~f_{\mathrm{obj}}(\vec{q})+\frac{\rho}{2}\norm{\vec{q}-\vec{z}^{k}+\frac{\vec{y}^{k}}{\rho}}_{2}^{2}\\
     \label{eqn:admmb}
     \vec{z}^{k+1}&=\underset{\vec{z}}\argmax ~\operatorname{g}_c(\vec{z})+\frac{\rho}{2}\norm{\vec{q}^{k+1}-\vec{z}+\frac{\vec{y}^{k}}{\rho}}_{2}^{2}\\
     \label{eqn:admmc}
     \vec{y}^{k+1}&=\vec{y}^k+\rho(\vec{q}^{k+1}-\vec{z}^{k+1}) \, .
     \end{align}%
    \end{subequations}%
The optimal condition is defined as~$\norm{\vec{q}^{k+1}-\vec{z}^{k+1}}_2\leq \epsilon^{\mathrm{primal}}$ and~$ \norm{\rho(\vec{z}^{k+1}-\vec{z}^k)}_2\leq \epsilon^{\mathrm{dual}}$, where~$\epsilon^{\mathrm{primal}}$ and~$\epsilon^{\mathrm{dual}}$ are two predefined optimality tolerances. Each sub problem is solved by using Bayesian optimization. The algorithm is summarized in Alg.~\ref{alg:ContactPointOptimization}.
 \begin{algorithm}[t]
     \small
     \begin{algorithmic}[1]
         \Require $f_{\mathrm{GPIS}}$, $\Transformation_{\mathrm{palm}}(\Rot,\vec{t}_{\mathrm{palm}})$
         \State Compute Chart $\mathcal{C}_i$ and get $\mathbf{\Phi}_{i}(\vec{t}_{\mathrm{palm}})$  solving equation~\ref{eqn:tangentSpace}
         \For{n = 0 to $n_{\mathrm{seed}}$ }
         \State $\vec{t}^{\prime}=\vec{t}_{\mathrm{palm}}+\mathbf{\Phi}_{i}({\vec{t}_{\mathrm{palm}}}) \mathbf{u}_{\mathrm{rand}}$
         \For{$\mathrm{iter}$ = 0 to $\mathrm{max}_{iter}$}
         \State Solve~$\vec{q}^{k+1}=\operatorname{BO}_q\bigl(f_{\mathrm{obj}}(\vec{q})+\frac{\rho}{2}\norm{\vec{q}-\vec{z}^{k}+\frac{\vec{y}^{k}}{\rho}}_{2}^{2}\bigr)$
         \State Solve~$\vec{z}^{k+1}=\operatorname{BO}_z\bigl(g_c(\vec{z})+\frac{\rho}{2}\norm{\vec{q}^{k+1}-\vec{z}+\frac{\vec{y}^{k}}{\rho}}_{2}^{2}\bigr)$
         \State Update~$\vec{y}^{k+1}=\vec{y}^k+\rho(\vec{q}^{k+1}-\vec{z}^{k+1})$
         \State Get \hbox{$\epsilon_1$ =$\norm{\vec{q}^{k+1}-\vec{z}^{k+1}}_2$}, {$\epsilon_2 =$ $\norm{\rho(\vec{z}^{k+1}-\vec{z}^k)}_2$}
         \If {${\epsilon_1\leq \epsilon^{\mathrm{primal}}}$ and ${\epsilon_2\leq \epsilon^{\mathrm{dual}}}$} break
         \EndIf
         \State $\vec{y}^{k}\leftarrow \vec{y}^{k+1},\vec{z}^{k} \leftarrow \vec{z}^{k+1} $
         \State Update $\operatorname{BO}_q$, $\operatorname{BO}_z$
         \EndFor 
         \EndFor
         \State \Return $\vec{q}^{k+1},\Transformation_{\mathrm{palm}}(\Rot,\vec{t}^{\prime})$
        \end{algorithmic} 
        \caption{ADMM-Based Contact Point Optimization~(ADMM-CP-Opt)}
        \label{alg:ContactPointOptimization}
\end{algorithm}
\subsection{Integration of HPP-Opt and ADMM-CP-Opt}
The HPP-Opt and ADMM-CP-Opt are in detail introduced in the previous section. This section will integrate both optimizations in one framework and aim to find a near-global optimized grasp. The integration process is summarized in Alg. ~(\ref{alg:BO-GPISAtlas}. Since HPP-Opt does not consider the hand finger configuration and the grasp is executed relying on the function of \emph{AutoGrasp} from Graspit!~~\cite{Miller2004GraspitAV},  we executed ADMM-CP-Opt by using the result of HPP-Opt. The final solution combines the palm pose from HPP-Opt and the hand finger from ADMM-CP-Opt. It can seem that the ADMM-CP-Opt hat at least the one solution as HPP-Opt.
\begin{algorithm}[t]
    \small
    \begin{algorithmic}[1]
        \Require $n_{\mathrm{dim}}$, $n_{\mathrm{iter}}$ 
        \State Get LHS \hbox{$\mathcal{D}_{0:t-1}=\bigl\lbrace\left(\mathbf{x}_{0}, y_{0}\right), \ldots,\left(\mathbf{x}_{t-1}, y_{t-1}\right)\bigr\rbrace$}
        \State Fit the Gaussian process Model $p\left(y | \mathbf{x}, \mathcal{D}_{0:t-1}\right)$
        \State Optimized hyper Parameter Rrop
        \For {$\mathrm{t}=1$ to $n_{\mathrm{iter}}$} 
        \State $\mathbf{x}_{\mathrm{palm},t}=\underset{{\mathbf{x}}}\argmax \operatorname{EI}\left(\mathbf{x} | \mathcal{D}_{0 : t-1}\right)$
        \State Find an no collision contact, and get $\vec{q},\vec{x}_{\mathrm{palm}}$
        \State $\mathcal{G}_1$=$\operatorname{GPISAtlas}$ ($f_{\mathrm{GPIS}}$, $\vec{x}_{\mathrm{palm}}$)
        \State $\{\mathcal{G}_{\mathrm{best}}(\vec{x}_{\mathrm{palm},t},\vec{q}),y_{\mathrm{max},t}\}$=$\underset{\mathcal{G}_1}\argmax f_{\mathrm{obj}}(\vec{q},\vec{x}_{\mathrm{palm}})$
        \State $\mathcal{G}_2(\vec{x}_{\mathrm{palm},t}^{\prime},\vec{q}^{\prime})$ = Admm-CP-Opt$\bigl(f_{\mathrm{GPIS}}, \Transformation_{\mathrm{palm}}(\vec{x}_{\mathrm{palm},t})\bigr)$
        \State $y_{\mathrm{max},t}^{\prime}=f_{\mathrm{obj}}(\vec{q}^{\prime},\vec{x}_{\mathrm{palm}}^{\prime})$
        \State Add new sample $\mathcal{D}_{0 : t}=\bigl\lbrace\mathcal{D}_{0 : t-1},(\vec{x}_{\mathrm{palm},t}^{\prime},y_{\mathrm{max},t}^{\prime})\bigr\rbrace$
        \State update Gaussian process model~(GP)
        \EndFor
        \State $\mathbf{x}_{\mathrm{palm}}^{+}=\underset{\mathbf{x}_{i} \in \mathbf{x}_{0 : t}}\argmax f_{\mathrm{obj}}\left(\mathbf{x}_{i}\right)$
        \State \Return $\mathbf{x}_{\mathrm{palm}}^{+}$
    \end{algorithmic} 
    \caption{Integration of HPP-Opt and ADMM-CP-Opt}
    \label{alg:BO-GPISAtlas}
\end{algorithm}

\section{Experiment}
\label{sec:experment}
\begin{figure}[tb]
    \centering
    \begin{subfigure}[b]{0.35\linewidth}
         \centering
            \includegraphics[width=1\linewidth]{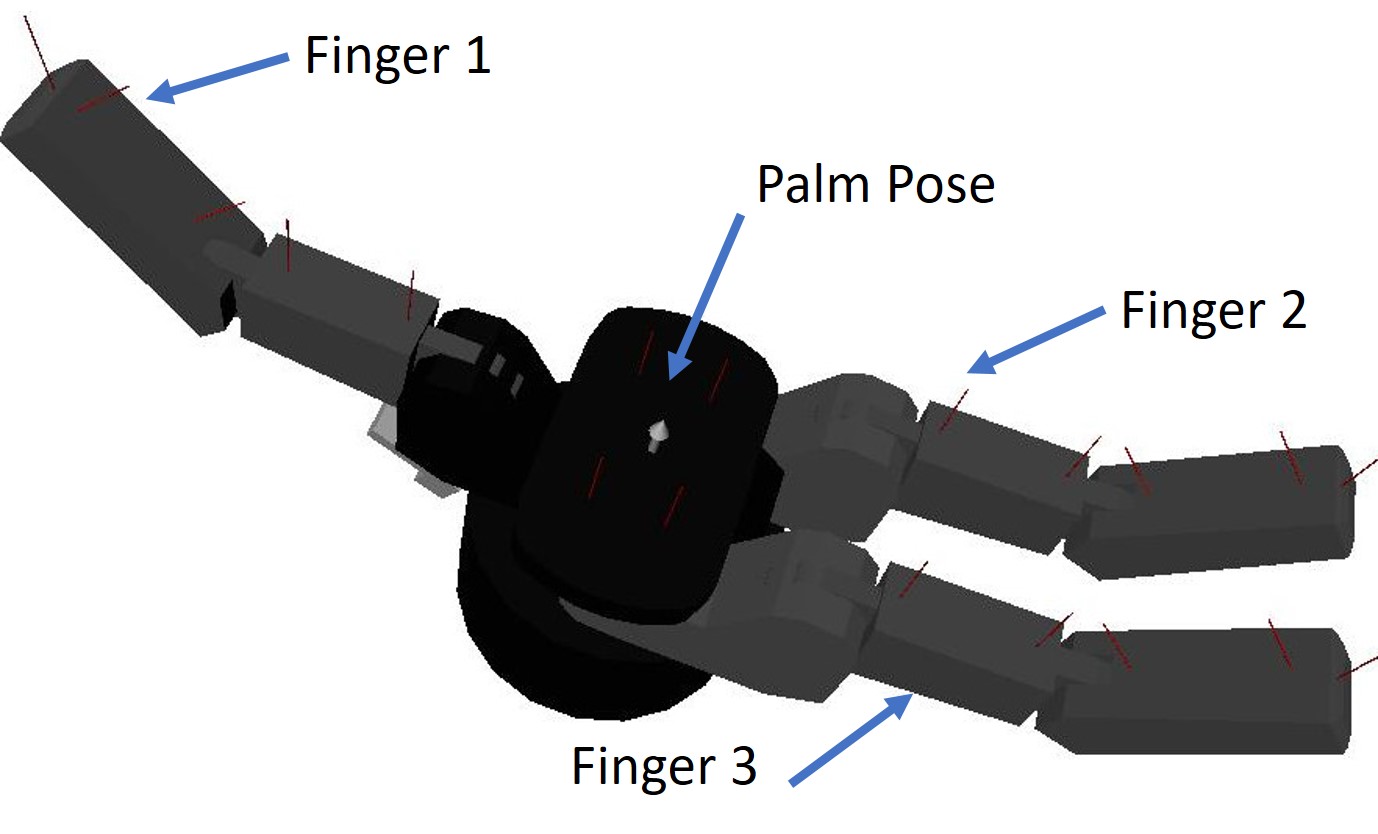}
            \caption{}%
         \label{fig:BarretHand}
    \end{subfigure}
    \begin{subfigure}[b]{0.20\linewidth}
         \centering
        \includegraphics[width=1\linewidth]{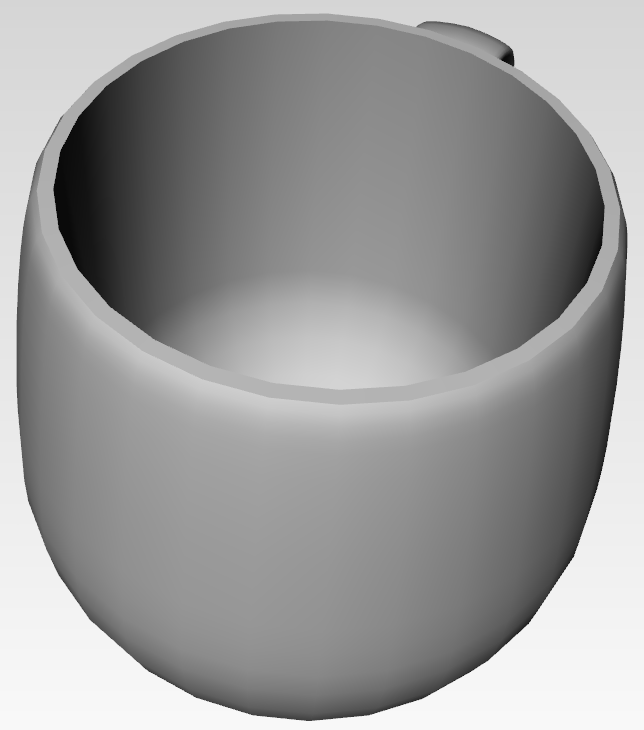}
        \caption{}%
        \label{fig:mug}
    \end{subfigure}
    \caption{Visualization of \subref{fig:BarretHand}~Barret hand and \subref{fig:mug}~Object mug.}
\end{figure}

Simulation results are introduced in this section to verify the effectiveness of our algorithm. The experiment is executed in the platform Graspit!~\cite{Miller2004GraspitAV} by using Barret hand, as shown in Fig.~\ref{fig:BarretHand}. 
Barret hand hat three fingers, finger one hat two joints, where the last joint is under-actuated. Finger two and three have one common joint. Each finger has three joints where both fingers have a passive joint. Therefore the Barret hand fingers have totaled 4 DOFs. The hand palm pose is denoted as a 6-dimensional vector. Therefore the whole Barret hand system has totaled 10 DOFs. The experiment's graspable object is stored in the mesh file, such as one example: mug in Fig.~{\ref{fig:mug}}. The GPIS describes the object by using the mesh triangle verities as the input. Our approach achieves a 95\% success rate on various commonly used objects with diverse appearances, scales, and weights compared to the other algorithm. All evaluations were performed on a laptop with a~\SI{2.6}{\giga\hertz} Intel Core i7-6700HQ and~\SI{16}{\giga\byte} of RAM.
\subsection{Experiment on HPP-Opt}
\label{sec:resutOfPP-OPT}
\begin{figure}[tb]
     \centering
     \newcommand\x{0.24}
     \begin{subfigure}[b]{\x\linewidth}%
         \centering
         \includegraphics[width=\linewidth] {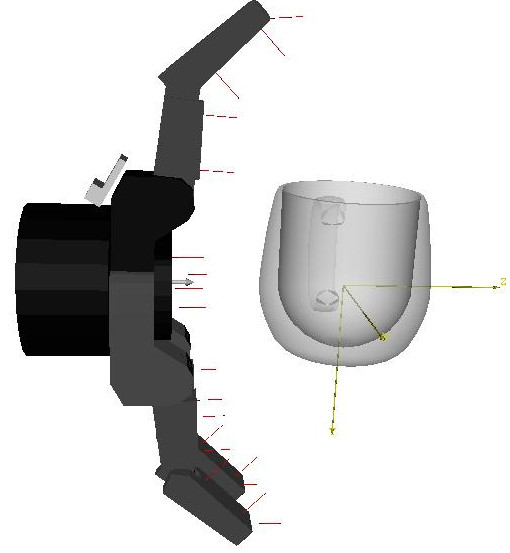}%
         \caption{Mug: Initial}
         \label{subfig:initiConfigurationBarretMug}
        \end{subfigure}%
        \hfill
        \begin{subfigure}[b]{\x\linewidth}%
            \centering
            \includegraphics[width=\linewidth] {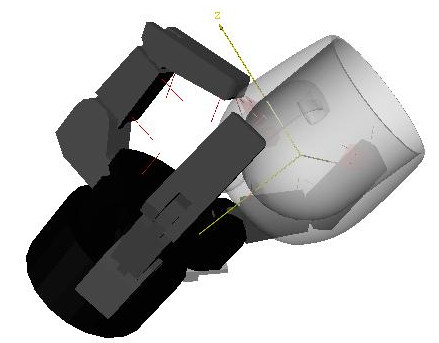}%
            \caption{Random Alg}
            \label{subfig:randomAlgBarretMug}
        \end{subfigure}%
        \hfill
        \begin{subfigure}[b]{\x\linewidth}%
            \centering
            \includegraphics[width=\linewidth] {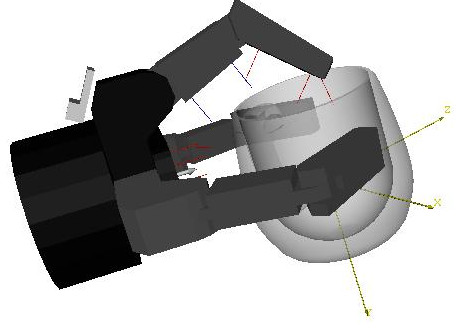}%
            \caption{EigenGrasp}
            \label{subfig:EigenAlgBarretMug}
        \end{subfigure}%
        \hfill
        \begin{subfigure}[b]{\x\linewidth}%
            \centering
            \includegraphics[width=\linewidth] {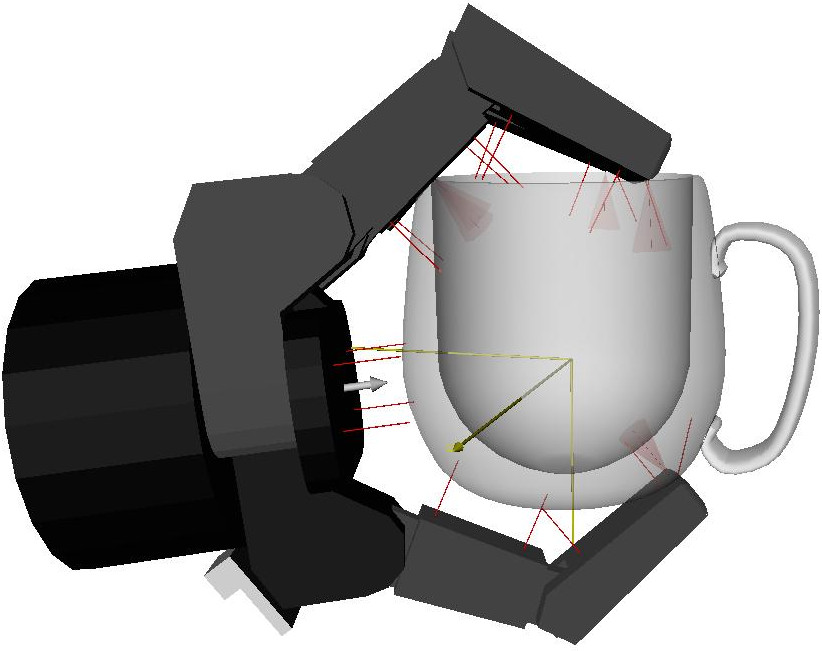}%
            \caption{HPP-Opt}
            \label{subfig:ppOPTAlgBarretMug}
        \end{subfigure}%
        \\\medskip%
        \begin{subfigure}[b]{\x\linewidth}%
            \centering
            \includegraphics[width=\linewidth] {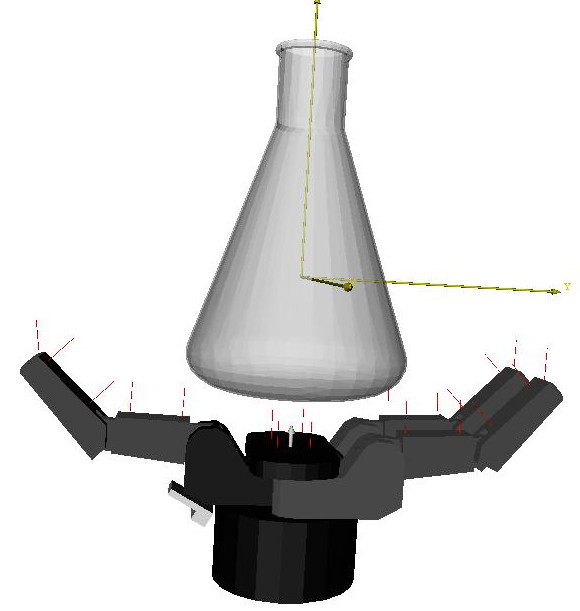}%
            \caption{Flask: Initial}
            \label{subfig:initiConfigurationBarretflask}
        \end{subfigure}%
        \hfill
        \begin{subfigure}[b]{\x\linewidth}%
            \centering
            \includegraphics[width=\linewidth] {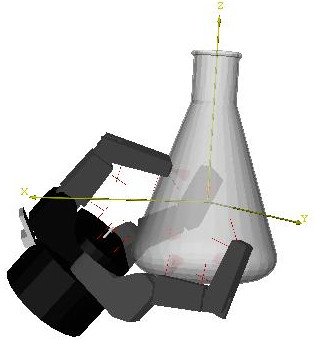}%
            \caption{Random Alg}
            \label{subfig:randomAlgBarretflask}
        \end{subfigure}%
        \hfill
        \begin{subfigure}[b]{\x\linewidth}%
            \centering
            \includegraphics[width=\linewidth] {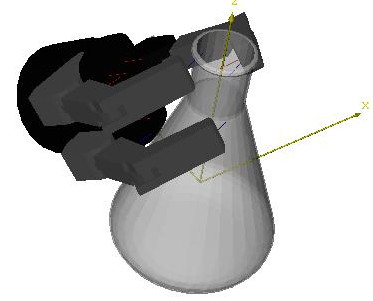}%
            \caption{EigenGrasp}
            \label{subfig:EigenAlgBarretflask}
        \end{subfigure}%
        \hfill
        \begin{subfigure}[b]{\x\linewidth}%
            \centering
            \includegraphics[width=\linewidth] {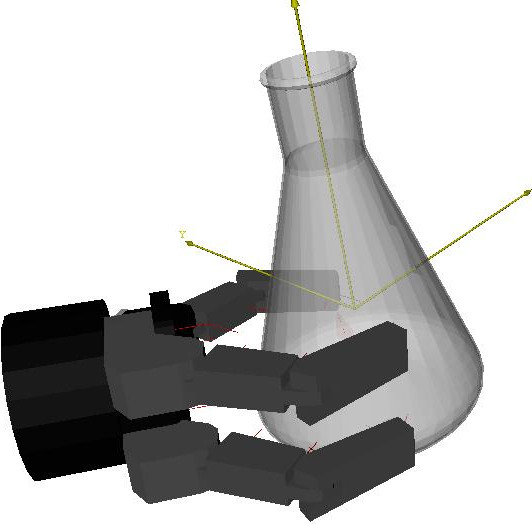}%
            \caption{HPP-Opt}
            \label{subfig:ppOPTAlgBarretflask}
        \end{subfigure}%
        \caption{ Comparison of different Grasp planning Algorithm examples for different objects with the multi-fingered hand}
        \label{fig:comparisonwithDifferentPlanner}
    \end{figure}

 In this section, the algorithm HPP-Opt is compared with other grasp planning. The first grasp planning approach is a simulated annealing grasp planner using an auto grasp quality energy as a search strategy, which behaves like a random grasp planning. The second approach uses an EigenGrasp planner combined with a simulated annealing solver to guide potential quality energy. The Bayesian optimization and simulated annealing both are global optimization solvers, where the latter one is a probabilistic technique for approximating the global optimum of a given function. We compare the algorithms with different kinds of shapes.  The experiment is conducted in a fixed time of $20$ seconds, and we average the first 20 best grasp candidates. The best grasp candidate of mug and flask from whole grasp candidates are visualized in Fig.~\ref{fig:comparisonwithDifferentPlanner}. 
The initial state of Barret hand and object are randomly defined, shown in Fig.~\ref{subfig:initiConfigurationBarretMug} and~\ref{subfig:initiConfigurationBarretflask}.  
It can be seen that the hand configuration selected by random grasp planning is skewed to the object, and contact points on the surface are not properly, and the resulting quality is also worse. The results of EigenGrasp show a better solution where the palm pose is trying to parallel to the object surface. The palm pose selected by HPP-Opt is more reasonable in comparison to the other two algorithms.  We show the quantitative result is in table~(\ref{tab:summary}). 
\begin{table}[tb]
    \caption{Evaluation of different grasp planning algorithm for all data sets. The results is an average of first 20 best Grasp Candidates. A greater value of epsilon and volume means a more stable grasp. The best result is highlighted in \emph{green}}
    \label{tab:summary}
    \centering%
     \begin{adjustbox}{width=\linewidth}%
        \sisetup{group-digits=false,table-number-alignment=right,table-parse-only=true,round-mode=places,round-precision=3,table-format=1.3}%
        \begin{tabular}{lcccccccc}%
            \toprule
            & \multicolumn{8}{c}{Algorithms}\\
			\cmidrule(r){2-9}
            & \multicolumn{2}{c}{random Alg}& \multicolumn{2}{c}{EigenGrasp}&\multicolumn{2}{c}{HPP-Opt}&\multicolumn{2}{c}{ADMM-CP-Opt}\\
             \cmidrule(r){2-3} \cmidrule(r){4-5} \cmidrule(r){6-7} \cmidrule(r){8-9} 
             quality & $q_\epsilon$&$q_{\mathrm{volume}}$& $q_\epsilon$&$q_{\mathrm{volume}}$& $q_\epsilon$&$q_{\mathrm{volume}}$& $q_\epsilon$&$q_{\mathrm{volume}}$\\
            \midrule
            Mug& 0& 9.5352e-05& 1.7281e-04&7.1981e-04&0.0555 &   0.0097& \color{best}0.06  &  \color{best}0.0161\\
            Flask& 0& 3.9603e-05& 4.9050e-04&2.0142e-04&\color{best}0.0107&    0.0014&\color{best} 0.0107 &  \color{best} 0.0026\\
            Phone& 0& 3.9209e-05& 0.0017&4.6942e-04&\color{best}0.0142   & 0.0035&\color{best}0.0142  &  \color{best}0.0042\\
            Sphere& 0.0042& 0.0019& 0&0.0011&0.0495  &  0.0121& \color{best}0.050  &  \color{best}0.0279\\
            Bishop& 0& 8.1935e-05& 0.0012&9.7708e-05&\color{best}0.0094 &   0.0011&\color{best}0.0094 &   \color{best}0.0016\\
            \bottomrule
        \end{tabular}%
       \end{adjustbox}
\end{table} 
In different kinds of geometry shapes, our algorithm can achieve a much more stable grasp than other algorithms. The epsilon quality achieved by the first approach is almost zero besides in the case of a sphere. The epsilon quality by Eigen grasps a minimal value. On average, the epsilon quality of HPP-Opt is 28.5 times greater than Eigen grasp's epsilon quality. And the HPP-Opt's volume quality is 32.1417 times greater than Eigen Grasp's volume quality. Furthermore, we show the best grasp candidate from the first 20 best grasp candidates in Table~\ref{tab:summary2}. 
\begin{table}[tb]
    \caption{The best grasp from 20 grasp candidates. The best result is highlighted in \emph{green}}
    \label{tab:summary2}
    \centering%
    \begin{adjustbox}{width=\linewidth}%
        \sisetup{group-digits=false,table-number-alignment=right,table-parse-only=true,round-mode=places,round-precision=3,table-format=1.3}%
        \begin{tabular}{lcccccccc}%
            \toprule
            & \multicolumn{8}{c}{Algorithms}\\
            \cmidrule(r){2-9}
            & \multicolumn{2}{c}{random Alg}& \multicolumn{2}{c}{EigenGrasp}&\multicolumn{2}{c}{HPP-opt}&\multicolumn{2}{c}{ADMM-CP-Opt}\\
            \cmidrule(r){2-3} \cmidrule(r){4-5} \cmidrule(r){6-7} \cmidrule(r){8-9} 
            quality & $q_\epsilon$&$q_{\mathrm{volume}}$& $q_\epsilon$&$q_{\mathrm{volume}}$& $q_\epsilon$&$q_{\mathrm{volume}}$& $q_\epsilon$&$q_{\mathrm{volume}}$\\
            \midrule
            Mug& 0 &   0.0012& 0.0065  &  0.0011&0.1039 &   0.0151& \color{best}0.1158  &  \color{best}0.0340\\
            Flask&0 &   0.0004& 0.0130 &   0.0031&0.0449&    0.0018&\color{best}0.0452&    \color{best}0.0019\\
            Phone& 0  &  0.0004& 0.0310  &  0.0007& 0.0208  &  \color{best}0.0198&\color{best}0.0335  &  0.0006\\
            Sphere& 0.0844  &  0.0301& 0&    0.0086&0.0914  &  0.0362& \color{best}0.1676  & \color{best} 0.0728\\
            Bishop& 0  &  0.0012& 0.0164  &  0.0004&0.0390 &   0.0020&\color{best}0.0444   & \color{best}0.0045\\
            \bottomrule
        \end{tabular}%

    \end{adjustbox}
\end{table}
\subsection{Experiment of integration HPP-Opt and ADMM-CP-Opt}
\begin{figure}[htbp]
    \centering
    \newcommand\x{0.3}
    \begin{subfigure}[b]{\x\linewidth}%
        \centering
        \includegraphics[width=\linewidth] {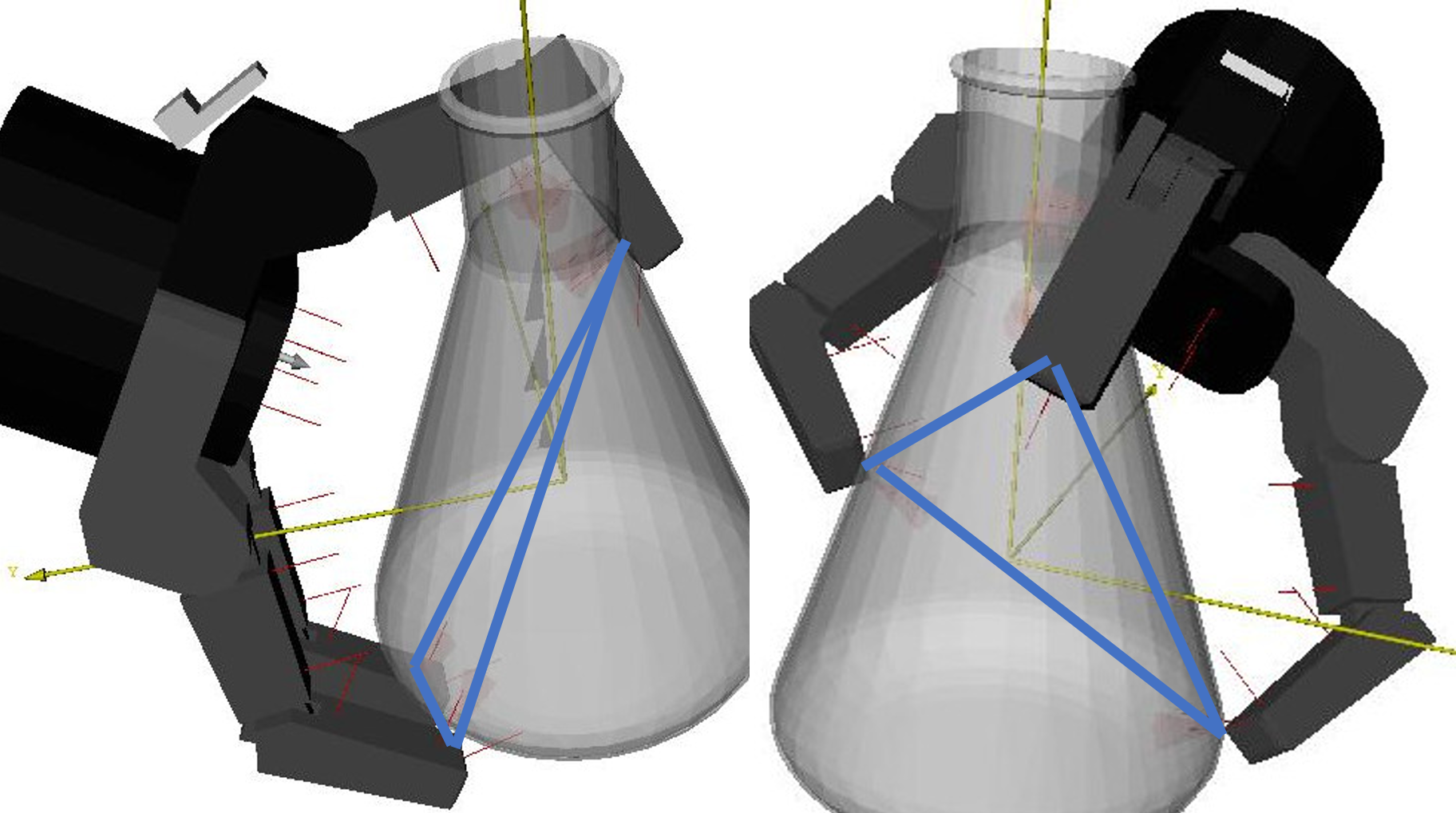}%
        \subcaption{}
        \label{subfig:ppOPT1}
    \end{subfigure}%
    \hfill
    \begin{subfigure}[b]{\x\linewidth}%
        \centering
        \includegraphics[width=\linewidth] {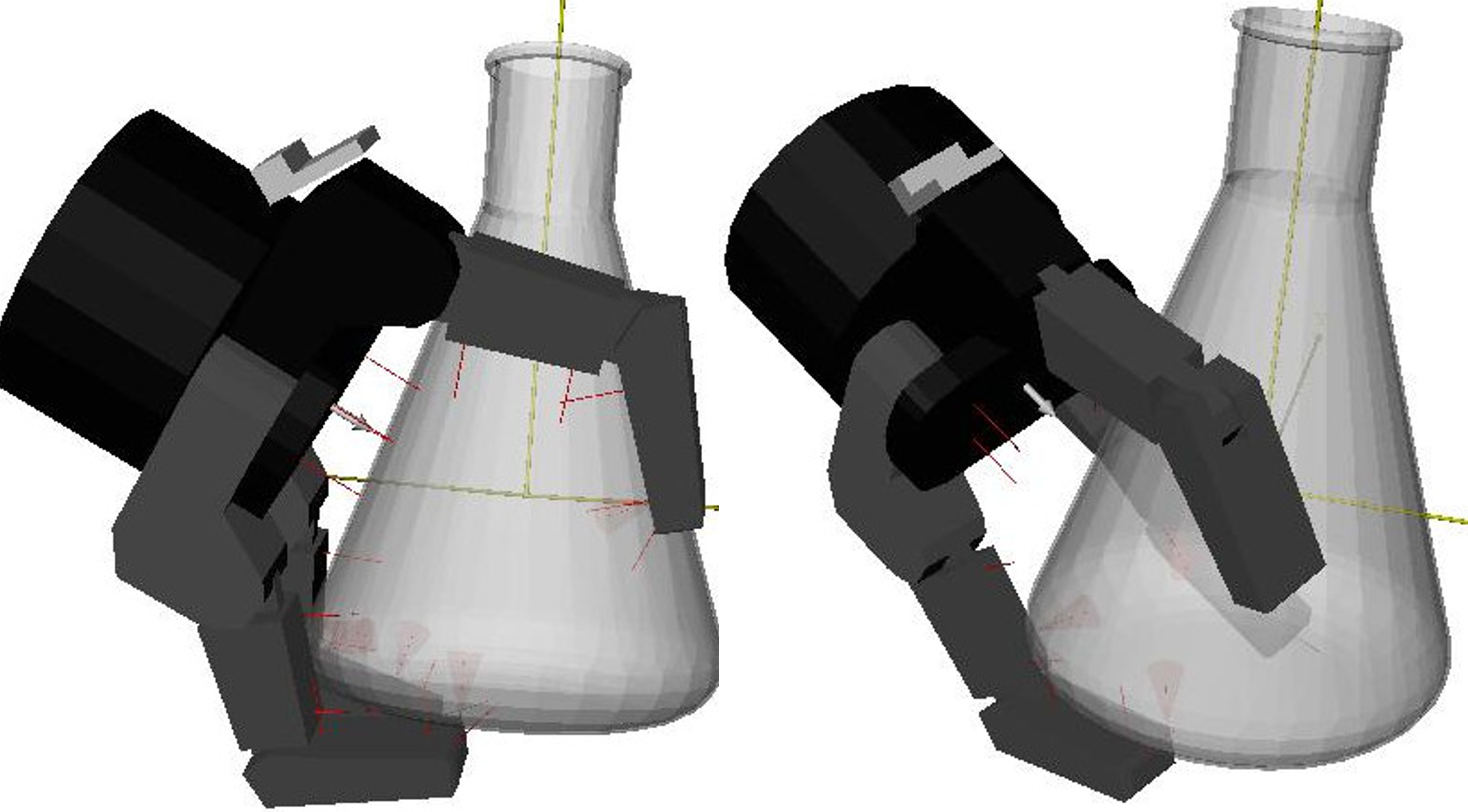}%
        \subcaption{}
        \label{subfig:cppOPT1}
    \end{subfigure}%
    \hfill
    \begin{subfigure}[b]{\x\linewidth}%
        \centering
        \includegraphics[width=\linewidth] {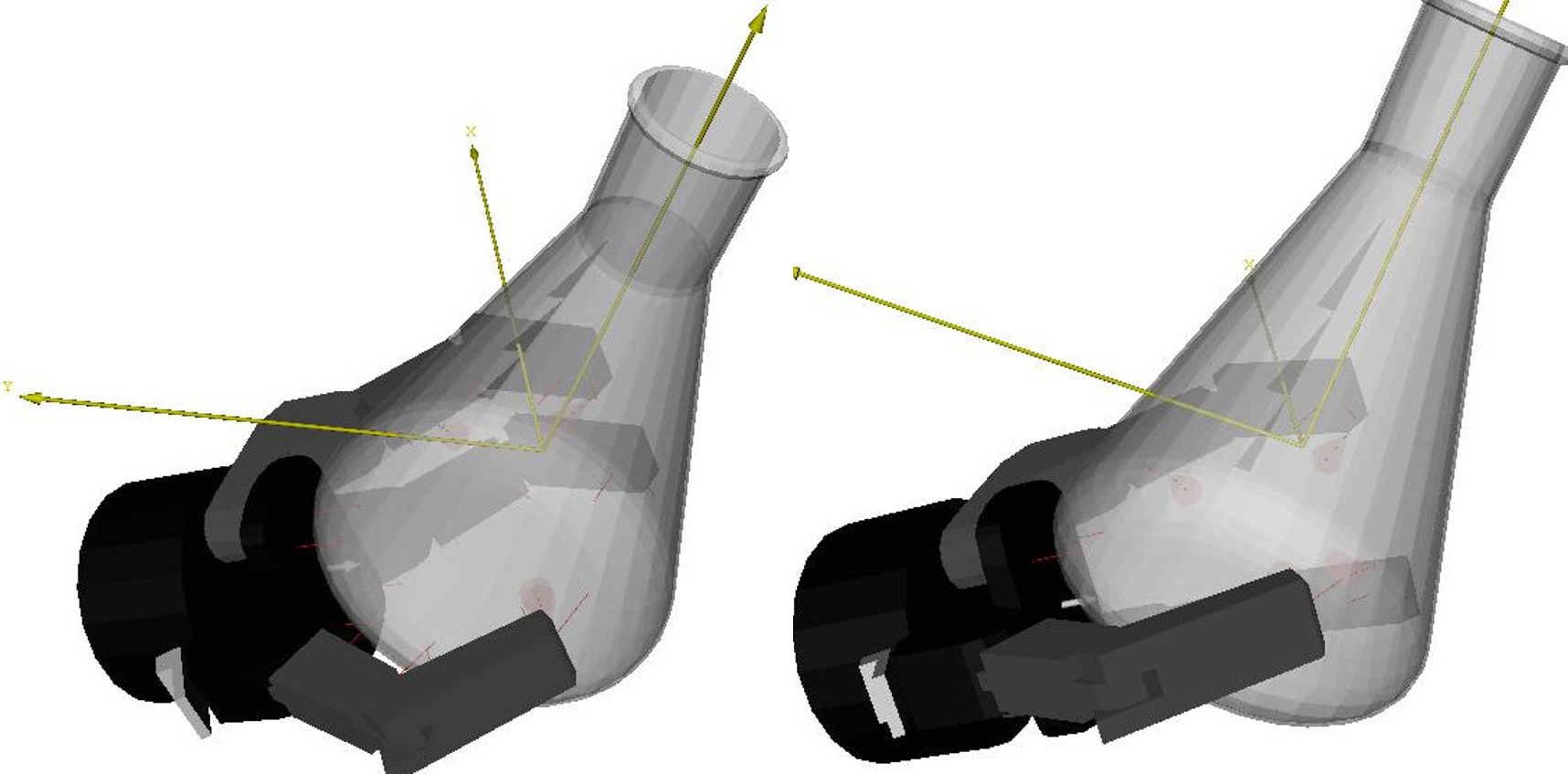}%
        \subcaption{}
        \label{subfig:cppOPT2}
    \end{subfigure}%
    \hfill
    \begin{subfigure}[b]{\x\linewidth}%
        \centering
        \includegraphics[width=\linewidth] {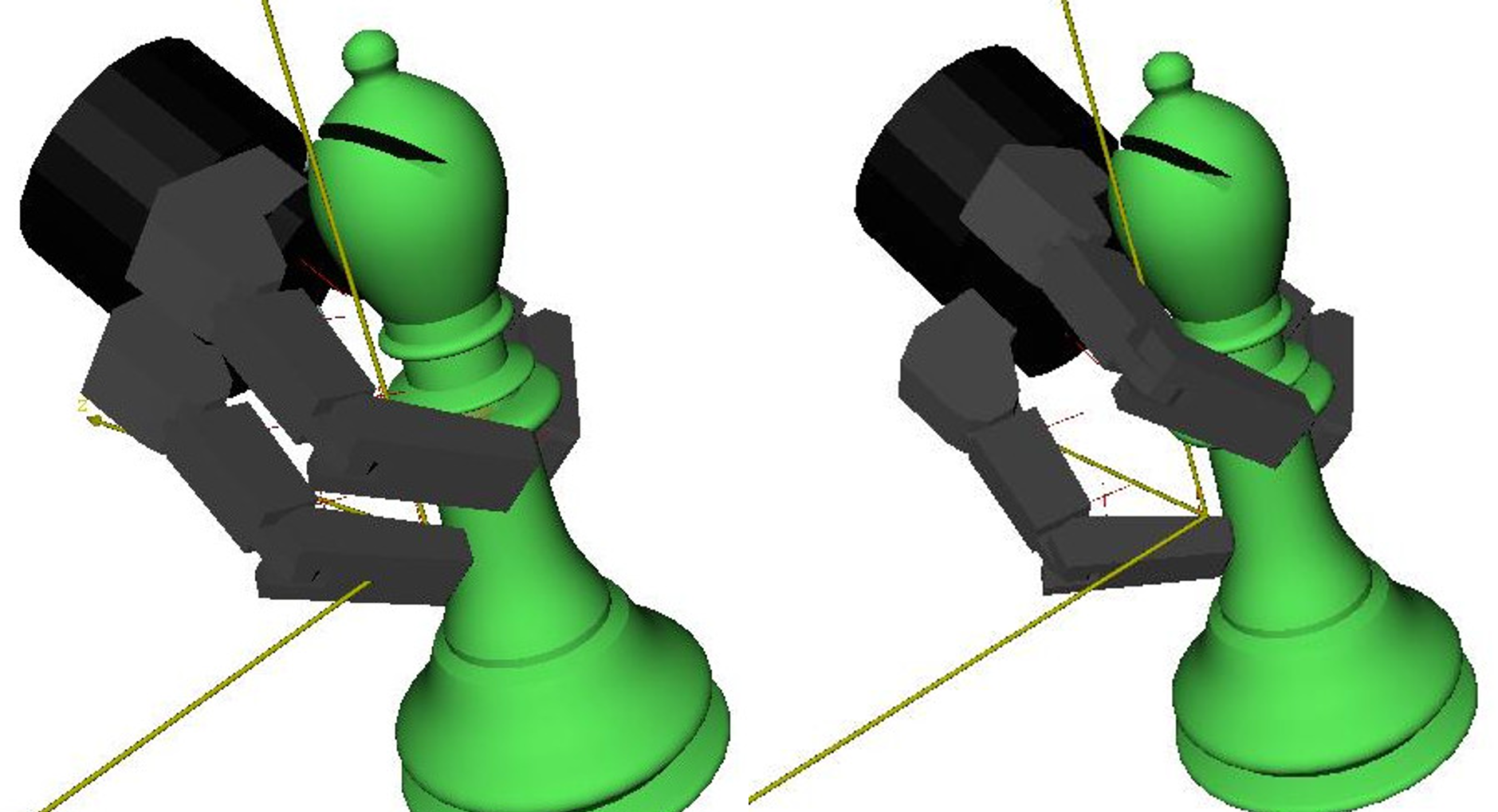}%
        \subcaption{}
        \label{subfig:cppOPT3}
    \end{subfigure}%
    \hfill
    \begin{subfigure}[b]{\x\linewidth}%
        \centering
        \includegraphics[width=\linewidth] {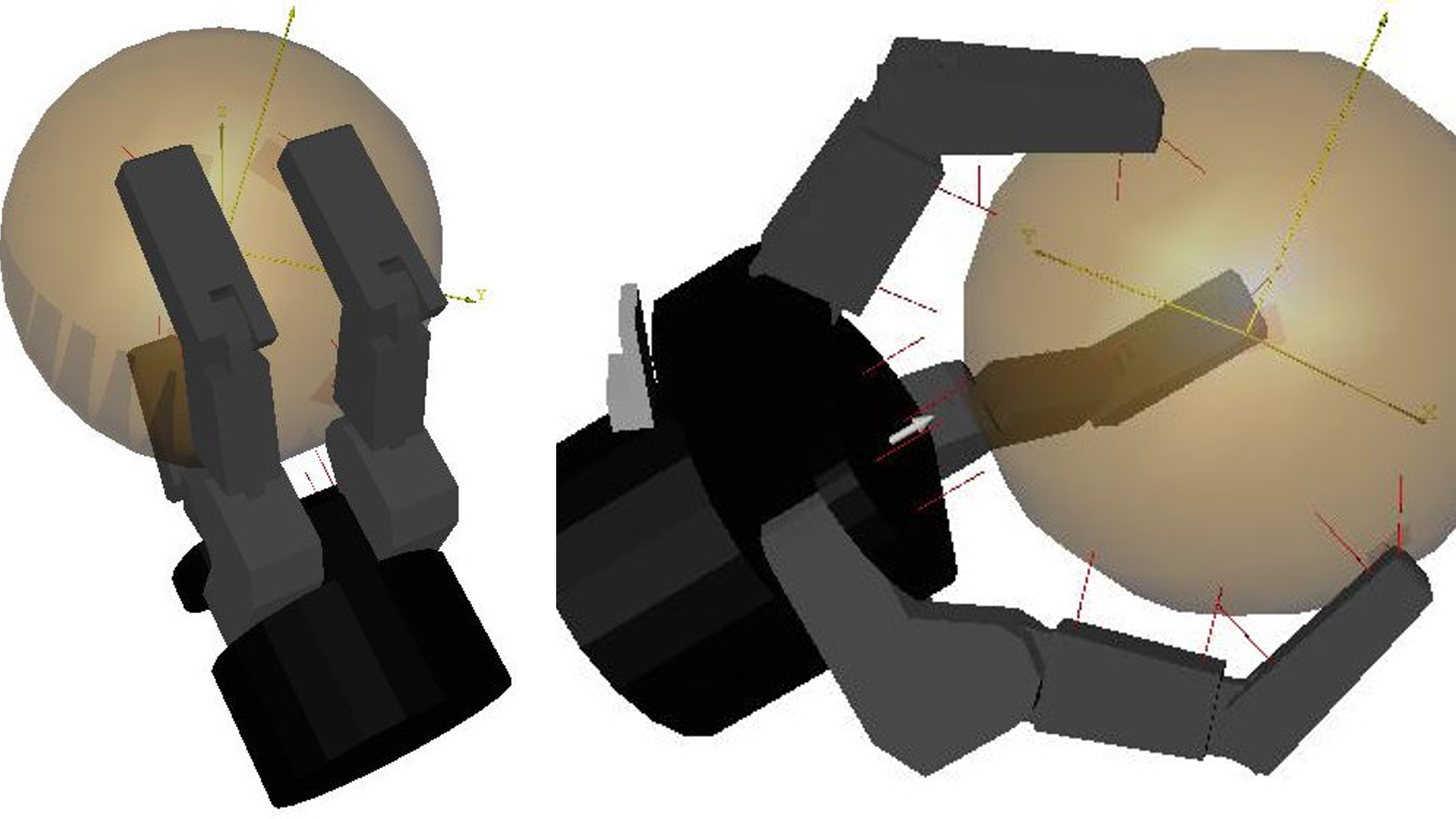}%
        \subcaption{}
        \label{subfig:cppOPT4}
    \end{subfigure}%
    \hfill
    \begin{subfigure}[b]{\x\linewidth}%
        \centering
        \includegraphics[width=\linewidth] {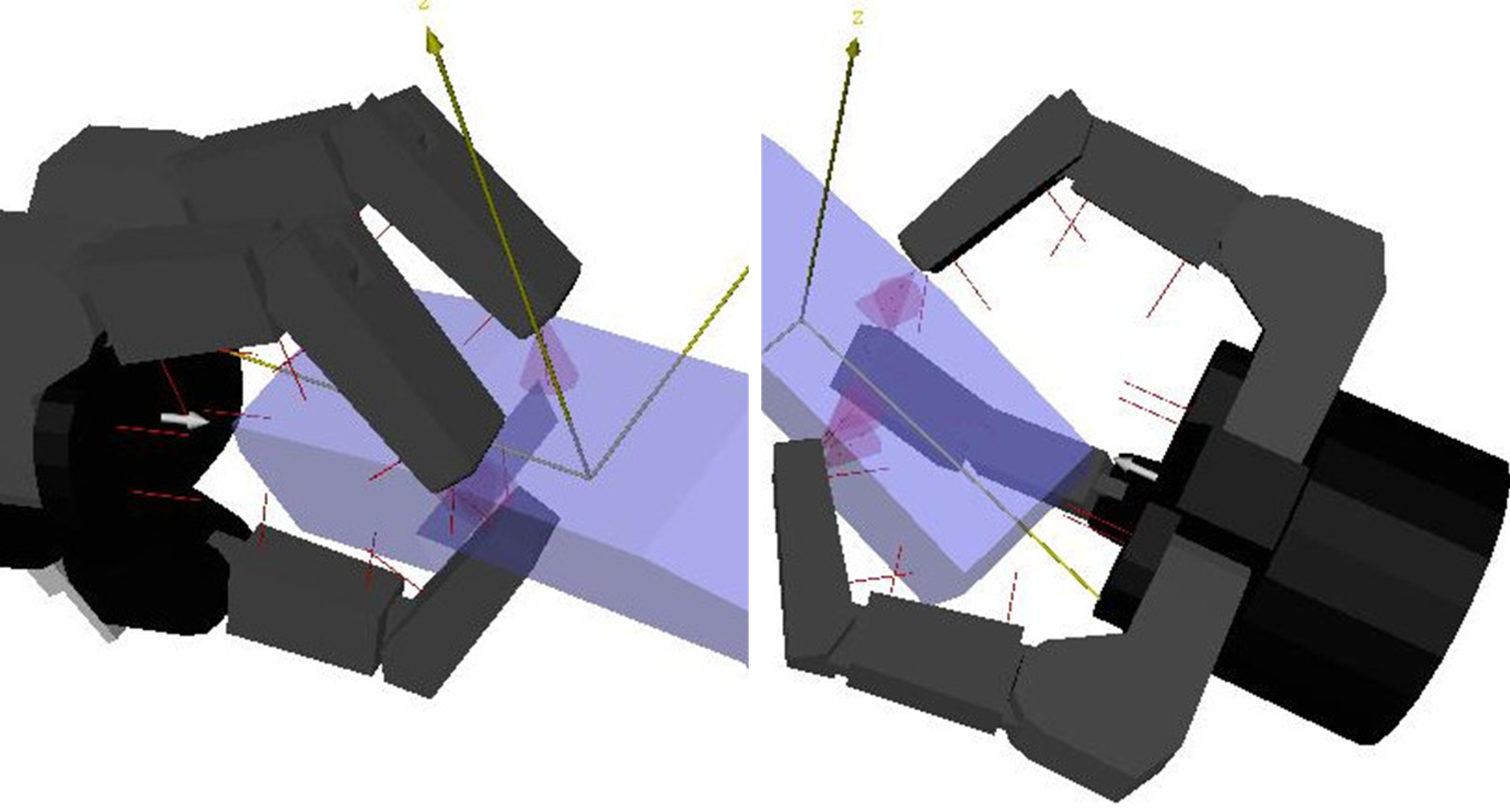}%
        \subcaption{}
        \label{subfig:cppOPT5}
    \end{subfigure}%
    \caption{The comparison result of HPP-Opt and ADMM-CP-Opt. In each sub figure, the left one is result of HPP-Opt, and the right one is improved by using ADMM-CP-Opt}
    \label{fig:pp-cp-Opt}
\end{figure}

In section~\ref{sec:resutOfPP-OPT}, hand palm pose is optimized based on the Bayesian optimization algorithm combing with local adaption, and hand finger configuration is set based on the function of \emph{AutoGrasp} from Graspit~\cite{Miller2004GraspitAV}. The principle of \emph{AutoGrasp}
is to close each hand finger DOF at a rate equal to a predefined speed factor multiple with its default velocity, and the movement is stopped at the contact point. Therefore ADMM-CP-Opt is used to assist the HPP-Opt to find a better hand finger configuration. The comparison result is visualized in Fig.~\ref{fig:pp-cp-Opt}. In the case ~(\ref{subfig:ppOPT1}), fingers two and three are too close and grasp the bottom of the flask. As a consequence, the resulting triangle has a small internal angle. Applying ADMM-CP-Opt splits the fingers two and three by maximizing epsilon and volume quality and makes the resulting triangle closer to the equilateral triangle with a  more stable grasp. The same improvement happen in the Fig.~\ref{subfig:cppOPT1} - Fig.~\ref{subfig:cppOPT5} as well. The solution founded by ADMM-CP-Opt is trying to make the resulting triangle as closer as the equilateral triangle. In Table~\ref{tab:summary}, ADMM-CP-Opt improves the volume quality of HPP-Opt. And In table
~\ref{tab:summary2}, ADMM-CP-Opt shows a better grasp than HPP-Opt in most cases under epsilon and volume quality metrics.

\section{Conclusion and Outlook}
\label{sec:conclusion}
We propose a new algorithm for grasp planning with multi-fingered hands by optimizing the hand palm pose and hand finger configuration separately. Using a global Bayesian optimization solver, no initial configuration is required, which shows superiority over the convex optimization solver. We propose a dual-stage optimization process by considering the independence of the hand palm pose and finger configuration. In the first stage, we utilize a GPIS to describe the graspable object so that collision checking of contact points can be integrated into the optimization framework. Furthermore, the chart on the object surface can be computed using the GPIS, which can be used to explore the local information of objects. Two ellipsoids are used to define the palm pose constraints domain. Relying on the first stage result, we apply an ADMM based Bayesian optimization to optimize the contact points. The whole process will switch between HPP-Opt and ADMM-CP-Opt. We collect the best 20 Grasps, and the final grasp is selected by ranking the grasp candidates under consideration of epsilon and volume quality. In this work, we describe only the object in GPIS. In future work, we can describe the hand part into GPIS so that the collision checking can be converted to a problem by querying the distance between two surfaces. Besides, the robot arm is not considered in the grasping scenario. It will be interesting to integrate the constraints of robot arm manipulability in the objective function.  

\bibliographystyle{IEEEtran}
\bibliography{IEEEabrv,myGraspPlanning}

\end{document}